\documentclass[Afour,sageh,times]{sagej}

\usepackage{moreverb,url}

\usepackage{graphics} 
\usepackage{epsfig} 
\usepackage{amsmath} 
\usepackage{amssymb}  
\usepackage{subcaption}
\usepackage{bm}
\usepackage{color}
\usepackage{epstopdf}

\usepackage[colorlinks,bookmarksopen,bookmarksnumbered,citecolor=red,urlcolor=red]{hyperref}

\newcommand\BibTeX{{\rmfamily B\kern-.05em \textsc{i\kern-.025em b}\kern-.08em
T\kern-.1667em\lower.7ex\hbox{E}\kern-.125emX}}

\def\volumeyear{2024}

\begin{document}

\runninghead{Ishihara et al.}

\title{Hierarchical Learning Framework for Whole-Body Model Predictive Control of a Real Humanoid Robot}

\author{Koji Ishihara\affilnum{1}, Hiroaki Gomi\affilnum{2} and Jun Morimoto\affilnum{1,3}}

\affiliation{\affilnum{1}Department of Brain Robot Interface, ATR Computational Neuroscience Laboratories, Kyoto, Japan\\
\affilnum{2}NTT Communication Science Laboratories, Nippon Telegraph and Telephone Co., Kanagawa, Japan\\
\affilnum{3}Graduate School of Informatics, Kyoto University, Kyoto, Japan}

\corrauth{Koji Ishihara, Department of Brain Robot Interface,
ATR Computational Neuroscience Laboratories,
2-2-2 Hikaridai,
Seika-cho, Soraku-gun,
Kyoto 619-0288, Japan.}

\email{ishihara-k@atr.jp}

\begin{abstract}
The simulation-to-real gap problem and the high computational burden of whole-body Model Predictive Control (whole-body MPC) continue to present challenges in generating a wide variety of movements using whole-body MPC for real humanoid robots. This paper presents a biologically-inspired hierarchical learning framework as a potential solution to the aforementioned problems. The proposed three-layer hierarchical framework enables the generation of multi-contact, dynamic behaviours even with low-frequency policy updates of whole-body MPC. The upper layer is responsible for learning an accurate dynamics model with the objective of reducing the discrepancy between the analytical model and the real system. This enables the computation of effective control policies using whole-body MPC. Subsequently, the middle and lower layers are tasked with learning additional policies to generate high-frequency control inputs.
In order to learn an accurate dynamics model in the upper layer, an augmented model using a deep residual network is trained by model-based reinforcement learning with stochastic whole-body MPC. The proposed framework was evaluated in 10 distinct motion learning scenarios, including jogging on a flat surface and skating on curved surfaces. The results demonstrate that a wide variety of motions can be successfully generated on a real humanoid robot using whole-body MPC through learning with the proposed framework.
\end{abstract}


\maketitle

\section{Introduction}
\label{sec: Introduction}
Model Predictive Control (MPC) is a method for controlling a system by iteratively reoptimizing and executing control policies, and is an active research topic in the field of legged robots such as humanoid robots \citep{katayama2023model}.
In particular, an MPC method that considers the whole-body dynamics of a robot is referred to as whole-body MPC.
Whole-body MPC has the potential to generate a variety of motions, as demonstrated on simulated humanoid robots \citep{tassa2012synthesis, erez2013integrated,ishihara2019full}.

Nevertheless, it remains challenging to generate diverse motions of a real humanoid robot using whole-body model MPC. The whole-body dynamics of a humanoid robot encompasses friction and contact phenomena that are difficult to model \citep{stewart2000rigid,fazeli2017empirical}. 
This prevents optimizing policies for a real robot using whole-body MPC.
Furthermore, given that a humanoid robot is a highly nonlinear, multi-degree-of-freedom system, optimization using its whole-body dynamics model requires a long computation time.
Feedback control with such slow policy updates is unable to mitigate the simulation-to-real gap, making it difficult to generate fast movements that dynamically interact with the environment.
For instance, \cite{koenemann2015whole} applied whole-body MPC to a real humanoid robot, but the derivation of policies took approximately 50 ms. This interval is insufficient for generating fast movements on a humanoid robot, as a policy update interval of 10 ms is necessary to generate behaviours without significantly slowing down the motion speed \citep{dantec2021whole}.

To address these issues, we propose a whole-body MPC framework inspired by human motor control. It has been suggesteed that control policies are reoptimized online in human motor control, and this process can be interpreted as MPC \citep{dimitriou2013temporal, nashed2014rapid, de2021online}.
An intriguing observation is that the time required to update the control policy for a voluntary response is considerably longer than for robot control, e.g., 150 - 180 ms \citep{de2021online}.
This leads to the question of the underlying mechanism that enables humans to generate a diverse range of motions in the presence of friction and contact despite these relatively slow policy updates.

Neuroscience studies has been suggested that accurate dynamic models can be acquired through the process of learning \citep{wolpert1998internal,shadmehr2010error}. 
The utilization of such precise models enables the prediction of precise states, which in turn facilitates the effectiveness of the voluntary response. However, a voluntary response is insufficient for dynamic interaction with the environment.
Consequently, hierarchical control with faster responses as well as the voluntary response is considered essential for motor control {\citep{gomi2008implicit, macpherson2021parallel}}. Long and short latency responses have been identified as fast responses, with reaction times ranging from 20 to 100 milliseconds \citep{pruszynski2012optimal}.

\begin{figure*}[t]
  \begin{center}
    \includegraphics[width=0.9\linewidth]{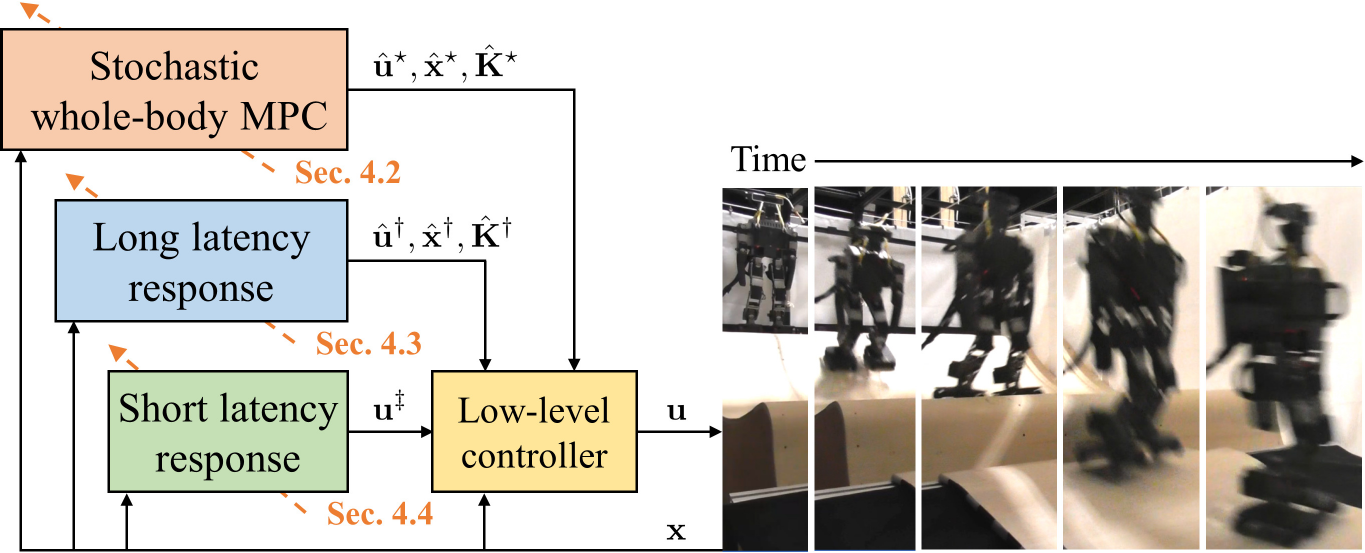}
    \caption{
      Overview of the proposed three-layer hierarchical learning framework:
      In the upper layer, an accurate dynamics model is learned and the control policy is computed by stochastic whole-body MPC.
      In the middle and lower layers, additional control policies are learned as long and short latency responses.
      The policies at each layer are sent to the low-level controller to generate the robot's motion.
    }
    \label{fig: overview}
  \end{center}
\end{figure*}

In light of the aforementioned studies, we propose a hierarchical learning framework in this paper. Figure \ref{fig: overview} provides an overview of the proposed framework. The framework comprises three layers of hierarchy: an upper layer of whole-body MPC, a middle layer of the long latency response, and a lower layer of the short latency response.
In the upper layer, an accurate dynamics model is acquired through model-based reinforcement learning (model-based RL) to bridge the simulation-to-real gap, thereby enabling the effective implementation of whole-body MPC. To learn the accurate dynamics model, we utilize an augmented dynamics model that combines an analytical dynamics model with a deep residual neural network. Furthermore, uncertainty is incorporated into the dynamics model to ensure stable learning in the upper layers through model-based RL.
Consequently, in the upper layer, policies are optimized by stochastic whole-body MPC that considers the uncertainty-aware dynamics. The middle and lower layers learn additional policies that allow the robot to generate movements despite the slow policy updates of the whole-body MPC. The control policies at each layer are integrated into a low-level controller, which generates the robot's motion.

The proposed framework was evaluated on a real humanoid robot in a task where the robot was required to learn 10 motions, including walking on a flat surface, jogging, and skating on a ramp and a jump ramp. Ablation studies were also conducted to validate the effectiveness of the proposed learning framework. The results demonstrated that our proposed whole-body MPC framework showed successful motion generation performances on a real humanoid robot.

The contributions of this paper are summarized as below:
\begin{enumerate} 
  \item 
  We developed a hierarchical learning framework that overcomes the simulation-to-real-gap problem and enables to generate robot motions even if whole-body MPC takes a long time to update its control policy.
  \item  
  We proposed to use an augmented model with a deep residual network and model-based RL with stochastic whole-body MPC to accurately and stably learn humanoid robot dynamics, and introduced biologically-inspired short- and long-latency responses to cope with the slow policy updates of whole-body MPC.
  \item 
  We showed that 10 different motions of a real humanoid robot could be generated by whole-body MPC through the learning with our proposed framework. 
\end{enumerate}

The rest of the paper is organized as follows:
Section \ref{sec: Related work} describes related works on whole-body MPC for humanoid robots.
Section \ref{sec: Preliminaries} gives preliminaries on dynamics models and whole-body MPC.
Section \ref{sec: Proposed framework} describes our proposed hierarchical learning framework.
Section \ref{sec: Experimental settings} provides the experimental settings and Section \ref{sec: Results} presents the experimental results.
Section \ref{sec: Conclusions} gives the conclusions of this paper and discusses future work.

\section{Related work}
\label{sec: Related work}
Reducing long computation time has been a major research topic for whole-body MPC.
\cite{tassa2012synthesis} reduced the computation time by simplifying the contact model and improving the line search, and demonstrated the potential of whole-body MPC to be applied to humanoid robots.
The whole-body MPC algorithm was applied to a real humanoid robot, and reaching motions were generated while maintaining balance, taking 50 ms for the MPC computation \citep{koenemann2015whole}.
\cite{li2021model} and our previous work \citep{ishihara2019full} developed whole-body MPC algorithms that reduce computation time using low-dimensional dynamics models.
We generated a slow skating motion on a flat ramp for a real humanoid robot, where the MPC control policy was updated every 20 ms \citep{ishihara2019full}.
\cite{dantec2022whole} and \cite{galliker2022planar} generated walking and climbing steps for real humanoid robots using efficient MPC algorithms that avoid the contact phase and timing optimization.
They achieved policy updates of whole-body MPC in 10 ms.
\cite{lembono2020learning} and \cite{dantec2021whole} constructed a motion memory to warm start a whole-body MPC optimizer with a good initial guess, and generated reaching motions for a real humanoid robot with policy updates in 10 ms \citep{dantec2021whole}.

In contrast, our approach in this paper is not aimed at reducing computation time.
Our hierarchical learning framework allows a humanoid robot to generate a variety of motions even when the policy updates of whole-body MPC are slow.
Furthermore, unlike the conventional whole-body MPC approaches for humanoid robots, our hierarchical learning framework bridges the gap between simulation and reality using model-based RL.
We demonstrated that the proposed three-layer hierarchical learning framework was able to generate a wider range of 10 types of motions on a real humanoid robot. The 10 motions include jogging and agile skating on a ramp and a jump ramp. A hierarchical control framework was proposed in our previous work \citep{ishihara2021computationally}, but it was a two-layer hierarchy and did not include dynamics model learning using model-based RL with stochastic whole-body MPC, and the variety of motions were generated only in simulation.

\section{Preliminaries}
\label{sec: Preliminaries}
\subsection{Analytical dynamics model}
Let us consider a humanoid robot with $n_{j}$ joints.
A dynamics model of the humanoid robot can be obtained analytically by modeling the robot as an underactuated tree of rigid bodies with a free-floating base.

We denote the position and orientation of the free-floating base as $\bm{q}_{1} \in \mathbb{R}^{7}$, the corresponding linear and angular velocities as $\bm{v}_{1} \in \mathbb{R}^{6}$, the joint angles as $\bm{q}_{2} \in \mathbb{R}^{n_{j}}$, the joint angular velocities as $\bm{v}_{2} \in \mathbb{R}^{n_{j}}$, and the joint torques as $\bm{\tau} \in \mathbb{R}^{n_{j}} $.
The equation of motion is given by
\begin{equation}
	\begin{bmatrix}
		\bm{M}_{11} & \bm{M}_{12} \\
		\bm{M}_{21} & \bm{M}_{22}
	\end{bmatrix}
	\begin{bmatrix}
		\dot{\bm{v}}_{1} \\
		\dot{\bm{v}}_{2}
	\end{bmatrix}
	+
	\begin{bmatrix}
		\bm{C}_{1}\\
		\bm{C}_{2}
	\end{bmatrix}
	= 
	\begin{bmatrix}
		\bm{0}\\
		\bm{\tau}
	\end{bmatrix}
	+
	\begin{bmatrix}
		\bm{J}_{1}^{\top}\\
		\bm{J}_{2}^{\top}
	\end{bmatrix}
	\bm{f}_{c},
	\label{eq: equation of motion}
\end{equation}		
where $\bm{M}_{**}$ is a component of the inertia matrix and $\bm{C}_{*}$ is a vector that includes the friction of each link and bias forces such as gravity, centrifugal force, and Coriolis force.
A component of the Jacobian matrix is $\bm{J}_{*}$ and the contact forces are denoted by $\bm{f}_{c}$.
At the $i$-th contact point, the contact force must lie inside the friction cone and the normal contact force must be non-negative.
Furthermore, the relative normal velocity at the contact point is also non-negative.
These constraints are written as
\begin{equation}
  \begin{array}{rcl}
	\sqrt{f_{cx,i}^{2}+f_{cy,i}^{2}} &\leq& \mu_{c} f_{cz,i},\\
  f_{cz,i} &\geq& 0, \\
  v_{cz,i} &\geq& 0,
  \end{array}
	\label{eq: contact constraint}
\end{equation}
where $f_{cx}$ and $f_{cy}$ are the tangent components of the contact force, $f_{cz}$ is the normal component, $v_{cz}$ is the relative normal velocity, and $\mu_{c}$ is the friction coefficient.

In this paper, we assume that the joint torques are determined by proportional control:
\begin{equation}
	\bm{\tau}=\bm{K}(\bm{q}_{2}^{r}-\bm{q}_{2}),
	\label{eq: proportional control}
\end{equation}
where $\bm{q}_{2}^{r}$ stands for the target joint angle and $\bm{K}$ is the proportional gain.

Discretizing (\ref{eq: equation of motion}) with a step size $\Delta t_{d}$, the analytical model of a humanoid robot can be obtained in the following form
\begin{equation}
	\textbf{x}_{k+1}=f_{\text{A}}(\textbf{x}_{k}, \textbf{u}_{k}, \theta_{\text{A}}),
  \label{eq: analytical dynamics}
\end{equation}
where $\textbf{x}=[\bm{q}_{1}^{\top}, \bm{q}_{2}^{\top}, \bm{v}_{1}^{\top}, \bm{v}_{2}^{\top}]^{\top}$ is the state vector, $\textbf{u}=\bm{q}_{2}^{r}$ is the control input and $k$ is a time index.
The analytical model has parameters $\theta_{\text{A}}$ such as mass, center of gravity location, and friction coefficient.
The parameters can be obtained from computer-aided design (CAD) drawings and extra data provided by robot manufacturers.

\subsection{Whole-body MPC}
MPC is a control method in which a control input sequence is obtained online, at each control period, by solving an optimal control problem.
The robot is controlled by repeatedly applying the first input of the optimal sequence in each control period.

Whole-body MPC refers to a class of MPC algorithms that find the control input sequence under the constraints of the whole-body dynamics model.
Let us define the state and control input sequences with an $N$-step horizon as $\textbf{X}_{k} = ( \textbf{x}_{k}, \textbf{x}_{k+1}, \cdots, \textbf{x}_{k+N-1} )$ and $\textbf{U}_{k} = ( \textbf{u}_{k}, \textbf{u}_{k+1}, \cdots, \textbf{u}_{k+N-2} )$.
The objective function is represented by the cumulative sum of the cost function $\ell$:
\begin{equation}
	J(\textbf{X}_{k},\textbf{U}_{k})=\sum_{t=k}^{k+N-2}\ell(\textbf{x}_{t},\textbf{u}_{t})+\ell(\textbf{x}_{k+N-1},\bm{0}).
	\label{eq: total cost}
\end{equation}
If the whole-body dynamics model is given by (\ref{eq: analytical dynamics}), the optimization problem of whole-body MPC is formulated as
\begin{equation}
	\begin{array}{cc}
	\min_{\textbf{U}_{k}} & J(\textbf{X}_{k},\textbf{U}_{k}) \\
	\text{s.t.} & \textbf{x}_{t+1}=f_{\text{A}}(\textbf{x}_{t},\textbf{u}_{t}, \theta_{\text{A}}) \\
  &g(\textbf{x}_{t},\textbf{u}_{t})\leq 0\\
	&(t=k \dots k+N-2),
	\label{eq: whole-body MPC}
	\end{array}
\end{equation}
where $\textbf{x}_{k}$ is the observed state of the robot or, if the observation is delayed, the predicted state $\hat{\textbf{x}}_{k}=f_{\text{A}}(\textbf{x}_{k-1},\textbf{u}_{k-1}, \theta_{\text{A}})$ from the delayed observation $\textbf{x}_{k-1}$ can also be used \citep{koenemann2015whole}.
The inequality constraints $g$ are defined by the contact constraints in (\ref{eq: contact constraint}).

In general, the optimal control problem for a nonlinear system such as a humanoid robot cannot be solved analytically.
Therefore, numerical optimization techniques are used to compute the optimal input sequence $\textbf{U}_{k}^{\star} = ( \textbf{u}_{k}^{\star}, \textbf{u}_{k+1}^{\star}, \cdots, \textbf{u}_{k+N-2}^{\star} )$.
Then, the input applied to the robot, $\textbf{u}_{k}$, is the first element of the optimal control sequence:
\begin{equation}
	\textbf{u}_{k} = \textbf{u}_{k}^{\star}.
	\label{eq: applied input}
\end{equation}
Alternatively, if the optimal state $\textbf{x}^{\star}$ and feedback gain $\textbf{K}^{\star}$ can be obtained as a byproduct of optimization, the following control policy can be used \citep{mitrovic2010optimal}:
\begin{equation}
	\textbf{u}_{k} = \textbf{u}_{k}^{\star}+\textbf{K}_{k}^{\star}(\textbf{x}_{k}-\textbf{x}_{k}^{\star}).
	\label{eq: applied input with feedback}
\end{equation}
Here, the subtraction (or addition) of the base orientation requires special care, such as the use of the boxminus (or boxplus) operator \citep{hertzberg2013integrating}.

\subsection{Issue of Whole-body MPC}
The numerical optimization, however, cannot compute effective control policies for the real robot due to the gap between the analytical model and the real robot.
This simulation-to-real gap occurs because model parameters obtained from CAD drawings can be inaccurate \citep{swevers2007dynamic}, and friction and contact are difficult to model accurately \citep{stewart2000rigid,fazeli2017empirical}.

Furthermore, the numerical optimization must solve a large-scale optimization problem because a humanoid robot has a large number of degrees of freedom.
The computational load is higher when iterative calculations are performed to accurately compute contact forces.
Thus, solving the numerical optimization takes a lot of computation time, and the interval between applying $\textbf{u}_{k}$ to $\textbf{u}_{k+1}$ to the robot becomes longer.
Such a frequency of policy updates is too slow to generate robot motion.

\section{Proposed framework}
\label{sec: Proposed framework}

To address these issues, we develop a biologically-inspired hierarchical learning framework in this paper.
This section describes the details of our proposed framework.

\subsection{Overview}
Our proposed framework has a three-layered hierarchical structure which consists of an upper layer, a middle layer, and a lower layer (see Figure \ref{fig: overview}).

The upper layer iteratively learns the whole-body dynamics model and improves the control policy of whole-body MPC in the manner of model-based RL to bridge the simulation-to-real gap.
The middle and lower layers learn additional control policies as long and short latency responses so that the robot's motion can be generated even if the MPC policy updates are slow.

The policies of each layer are combined into a control input $\textbf{u}$ in the low-level controller, and the control input is applied to the robot.
Specifically, the low-level controller receives the optimal input $\textbf{u}^{\star}$, state $\textbf{x}^{\star}$, and gain $\textbf{K}^{\star}$ of whole-body MPC from the upper layer, the learned input $\textbf{u}^{\dag}$, state $\textbf{x}^{\dag}$, and gain $\textbf{K}^{\dag}$ from the middle layer, and the input $\textbf{u}^{\ddag}$ from the lower layer.
The control input $\textbf{u}$ is computed as follows.
\begin{equation}
\begin{array}{rcl}
  \textbf{u}_{k}&=& \textbf{u}_{k}^{\star}+\textbf{K}_{k}^{\star}(\textbf{x}_{k}-\textbf{x}_{k}^{\star})\\ 
                &&+ \textbf{u}_{k}^{\dag}+\textbf{K}_{k}^{\dag}(\textbf{x}_{k}-\textbf{x}_{k}^{\dag}) \\
                &&+ \textbf{u}_{k}^{\ddag}.
	\label{eq: u act}
\end{array}
\end{equation}
Note that we used first-order hold to interpolate the inputs, states, and gains of the upper and middle layers so that smooth control inputs can be computed in the low-level controller.

\subsection{Upper layer: model-based RL with stochastic whole-body MPC}
The learning at the upper layers is performed in a model-based RL fashion, where the dynamics model and the MPC control policy are iteratively learned and refined.

Suppose we have an initial dynamics model and collect a motion dataset $\mathcal{D} = (\textbf{x}_{0}, \textbf{u}_{0}, \textbf{x}_{1}, \textbf{u}_{1}, \textbf{x}_{2}, \textbf{u}_{2}, \cdots )$.
First, the parameters of the analytical model (\ref{eq: analytical dynamics}) are identified from the data.
The training dataset of the analytical model is constructed as $N_{\text{A}}$ trajectories of states and inputs: $\mathcal{D}_{\text{A}} = (\textbf{X}_{0}, \textbf{U}_{0}, \cdots, \textbf{X}_{N_{\text{A}}-1}, \textbf{U}_{N_{\text{A}}-1} )$.
The parameters $\theta_{\text{A}}$ are estimated by minimizing the following $N$-step loss function around the initial parameters while ensuring physical consistency of the parameters
\begin{equation}
  E(\theta_{\text{A}})= \frac{1}{N_{\text{A}} N}\sum_{k=0}^{N_{\text{A}}-1} \sum_{t=1}^{N}\| \textbf{x}_{k+t}-\hat{\textbf{x}}_{k+t} \|_{W_{\text{A}}}^{2},
  \label{eq: analytical model objective function}
\end{equation}
where $\hat{\textbf{x}}$ is the predicted state which is computed as $\hat{\textbf{x}}_{k+t}=f_{\text{A}}(\hat{\textbf{x}}_{k+t-1}, \textbf{u}_{k+t-1}, \theta_{\text{A}})$ and $\hat{\textbf{x}}_{k}=\textbf{x}_{k}$.
The weight on the prediction error is denoted by $W_{\text{A}}$.

Then, the analytical model is augmented with a deterministic deep residual model $f_{\text{NN}}$:
\begin{eqnarray}
	\textbf{x}_{k+1}&=&f_{\text{A}}(\textbf{x}_{k}, \textbf{u}_{k}, \theta_{\text{A}})+f_{\text{NN}}(\textbf{x}_{k}, \textbf{u}_{k}, \theta_{\text{NN}}) \nonumber \\ 
  &=& f(\textbf{x}_{k}, \textbf{u}_{k}).
	\label{eq: residual model}
\end{eqnarray}
The parameters $\theta_{\text{NN}}$ of the residual model are learned by minimizing the following 1-step mean squared error 
\begin{equation}
	E(\theta_{\text{NN}})=\frac{1}{N_{\text{NN}}}\sum_{k=0}^{N_{\text{NN}}-1} \| \textbf{x}_{k+1}-\hat{\textbf{x}}_{k+1} \|^{2},
	\label{eq: residual model objective function}
\end{equation}
with $L_{2}$ reguralization.
The predicted state is computed as $\hat{\textbf{x}}_{k+1}=f(\textbf{x}_{k}, \textbf{u}_{k})$.
The training dataset of the residual model is constructed as $N_{\text{NN}}$ tuples of states and inputs: $\mathcal{D}_{\text{NN}} = ( (\textbf{x}_{0}, \textbf{u}_{0}, \textbf{x}_{1}), \cdots, (\textbf{x}_{N_{\text{NN}}-2}, \textbf{u}_{N_{\text{NN}}-2}, \textbf{x}_{N_{\text{NN}}-1}) )$.

Finally, the control policy is recomputed using the learned model and new motion data is collected.
The collected data are added to the motion dataset $\mathcal{D}$, and both the analytical and residual models are learned again.
This process is repeated until the maximum number of iterations $N_{\text{RL}}$ is reached.

However, learning is often unstable and suboptimal due to overfitting of the dynamics model to a small number of data samples, especially in the early learning phases.
One approach to alleviate this model bias problem is to incorporate uncertainty in the dynamics model \citep{deisenroth2011pilco}.
To do this, a stochastic term is added to (\ref{eq: residual model}):
\begin{equation}
	\textbf{x}_{k+1}=f(\textbf{x}_{k}, \textbf{u}_{k})+\sigma F(\textbf{x}_{k}, \textbf{u}_{k})\omega_{k},
	\label{eq: stochastic whole-body dynamics}
\end{equation}
where $\omega$ is zero-mean white Gaussian noise with unit covariance, $\sigma$ is the noise amplitude.
In this paper, we assume that the upper layer input has an uncertainty due to the model overfitting.
We choose to capture the uncertainty as a control-dependent noise that is multiplicative in the control input \citep{todorov2005generalized}.
To reflect the effect of this noise in the model, we approximate the function $F$ so that it maps the noise to the state depending on the magnitude of the upper layer input:
\begin{equation}
  F(\textbf{x}_{k}, \textbf{u}_{k})=
	(f(\textbf{x}_{k}, \textbf{u}_{k})- f(\textbf{x}_{k}, \textbf{u}_{k}^{\text{w/o up}}))/\sqrt{\Delta t_{d}}.
	\label{eq: input dependent noise}
\end{equation}
where $\textbf{u}^{\text{w/o up}}=\textbf{u}_{k}^{\dag}+\textbf{K}_{k}^{\dag}(\textbf{x}_{k}-\textbf{x}_{k}^{\dag})+\textbf{u}_{k}^{\ddag}$.

The control policiy is recomputed by stochastic whole-body MPC that minimizes the expectation of the objective function (\ref{eq: total cost}), subject to the constraints of the stochastic dynamics model (\ref{eq: stochastic whole-body dynamics}):
\begin{equation}
	\begin{array}{cc}
	\min_{\textbf{U}_{k}} & \mathbb{E}[J(\textbf{X}_{k},\textbf{U}_{k})] \\
	\text{s.t.} &\textbf{x}_{k}=\hat{\textbf{x}}_{k} \\
  & \textbf{x}_{t+1}=f(\textbf{x}_{t},\textbf{u}_{t})+\sigma F(\textbf{x}_{t}, \textbf{u}_{t})\omega_{t} \\
  &g(\textbf{x}_{t},\textbf{u}_{t})\leq 0\\
	&(t=k \dots k+N-2),
	\label{eq: stochastic whole-body MPC}
	\end{array}
\end{equation}
where $\hat{\textbf{x}}_{k}$ is a predicted state computed from the delayed observed state $\textbf{x}_{k-1}$ by $\hat{\textbf{x}}_{k}=f(\textbf{x}_{k-1}, \textbf{u}_{k-1})$.

The motion data is generated by applying the control input $\textbf{u}$ in (\ref{eq: u act}) to the robot.
To compute the control input, the optimized input $\textbf{u}^{\star}_{k}$, state $\textbf{x}^{\star}_{k}$, and gain $\textbf{K}^{\star}_{k}$ are sent from the upper layer to the low-level controller.

\subsection{Middle layer: learning of long latency response}
While the upper layer iteratively learns the dynamics model, the control policies of the middle and lower layers are also iteratively learned.

In human motor control, the voluntary response requires a long reaction time ($>100$ ms), and such a slow response alone is not sufficient for dynamic interaction with the environment.
Therefore, hierarchical control using fast responses such as long and short latency responses is considered essential for motion control \citep{gomi2008implicit,macpherson2021parallel}.
From this suggestion, we introduce such a hierarchical structure for humanoid robot control so that robot motions can be generated despite the slow policy updates of whole-body MPC.

Although the functional role of the long latency response is controversial, in this paper we implement the long latency response based on the hypothesis that it mimics the voluntary response \citep{hasan2005human}.
It avoids deviations from the unperturbed pattern of voluntary movements and affords sufficient time for the voluntary response to respond to the perturbation.

In our framework, the middle layer control policy is learned to mimic the upper layer control policy in an iterative learning control fashion.
Let $(\textbf{u}^{\star}_{0,i}, \textbf{u}^{\star}_{1,i}, \cdots, \textbf{u}^{\dag}_{0,i}, \textbf{u}^{\dag}_{1,i},\cdots )$ be the upper and middle layer inputs obtained at learning iteration $i$, then the middle layer inputs in the next learning iteration $i+1$ are yielded as follows
\begin{equation}
	\textbf{u}^{\dag}_{k,i+1} \leftarrow \textbf{u}^{\dag}_{k,i} + \alpha \textbf{u}^{\star}_{k,i} -\mu \textbf{u}^{\dag}_{k,i},
	\label{eq: u middle learn}
\end{equation}
where $\alpha$ and $\mu$ are the learning and forgetting rates, respectively.
In this learning rule, essentially, the middle layer input $\textbf{u}^{\dag}$ is updated in the same direction as the upper layer input $\textbf{u}^{\star}$ by the second term on the right-hand side.
Learning stops when the middle layer input completely mimics the upper layer input, i.e., the middle layer input acts like the upper layer input and the value of the upper layer input is zero.
The third term on the right-hand side is introduced to prevent learning from diverging, and its effect is small because the forgetting rate is a small parameter.
The middle layer gain $\textbf{K}^{\dag}$ is also obtained by the same learning rule.
The middle layer state $\textbf{x}^{\dag}$ is updated to be close to the upper layer state $\textbf{x}^{\star}$ with the following learning rule
\begin{equation}
	\textbf{x}^{\dag}_{k,i+1} \leftarrow \textbf{x}^{\dag}_{k,i} + \alpha (\textbf{x}^{\star}_{k,i}-\textbf{x}^{\dag}_{k,i}) -\mu (\textbf{x}^{\dag}_{k,i}-\textbf{x}^{\text{init}}_{k,i}),
	\label{eq: x middle learn}
\end{equation}
where $\textbf{x}^{\text{init}}$ is the initial value of the middle layer state.

The learned input $\textbf{u}^{\dag}_{k,i+1}$, state $\textbf{x}^{\dag}_{k,i+1}$, and gain $\textbf{K}^{\dag}_{k,i+1}$ are sent to the lower-level controller to compute the control input $\textbf{u}$ in (\ref{eq: u act}) at the next learning iteration $i+1$.

\subsection{Lower layer: learning of short latency response}
As the short latency response, we implement a pattern generator that produces primitive movements.
It is suggested that these lower-layer primitives, built in by evolution, allow the upper motor hierarchy to focus more on task-critical modifications and better optimize motor commands \citep{giszter2010spinal}.
To learn primitive motion patterns, we use Dynamic Movement Primitives (DMP) \citep{ijspeert2013dynamical}.
A primitive motion pattern is represented by a time series of $\textbf{x}^{\text{DMP}}=[\bm{q}_{1}^{\text{DMP}\top},\bm{q}_{2}^{\text{DMP}\top},\bm{v}_{1}^{\text{DMP}\top},\bm{v}_{2}^{\text{DMP}\top}]^{\top}$, where $\bm{q}_{1}^{\text{DMP}}$ is the position and orientation of the base link, $\bm{v}_{1}^{\text{DMP}}$ is the corresponding linear and angular velocities, $\bm{q}_{2}^{\text{DMP}}$ is the joint angles, and $\bm{v}_{2}^{\text{DMP}}$ is the joint angular velocities.
The pattern of the base link position and joint angles, for each degree of freedom, is learned as the following DMP trajectory $y$
\begin{equation}
    \begin{array}{rcl}
    \tau\dot{z}&=&\alpha_{z}(\beta_{z}(g-y)-z)+f, \\ 
    \tau\dot{y}&=&z,
	\label{eq: dmp}
    \end{array}
\end{equation}
where $\tau$ is the time constant, $\alpha_{z}$ and $\beta_{z}$ are positive constants, $g$ is the desired position on the motion pattern, and $f$ is the forcing term.
The corresponding velocity pattern for each degree of freedom is represented by $\dot{y}$.
On the other hand, the patterns of the base link orientation are learned as trajectories $\bm{q}$ of the quaternion-based DMP \citep{ude2014orientation}:
\begin{equation}
    \begin{array}{rcl}
    \tau\dot{\bm{\eta}}&=&\alpha_{z}(\beta_{z}(\bm{g}_{o}-\bm{q})-\bm{\eta})+\bm{f}_{o}, \\ 
    \tau\dot{\bm{q}}&=&\frac{1}{2}\bm{\eta}\ast\bm{q},
	\label{eq: quat dmp}
    \end{array}
\end{equation}
where $\bm{g}_{o}$ is the desired quantanion orientation, $\bm{f}_{o}$ is the forcing term, $\ast$ is the quaternion product, and $\bm{g}_{o}-\bm{q}$ is computed using the boxminus operator \citep{hertzberg2013integrating}.
The corresponding angular velocity patterns are represented using the trajectory $\bm{\eta}$.
The desired position and angles $g$ and orientation $\bm{g}_{o}$ are iteratively modified using the robot's state when the robot fails to generate motion, inspired by the suggestion of failure-driven motion adaptation by humans \citep{ikegami2021hierarchical}. 

Additionally, a stretch reflex are also implemented as the short latency response.
The reflex is activated when the joint angle approaches its limit and moves the angle in the opposite direction from the limit.
We design the reflex input by
\begin{equation}
	\textbf{u}^{\text{reflex}}_{k}=\textbf{K}^{\text{reflex}}(\zeta_{a}(\bm{q}_{2,k}^{\text{max}}-\bm{q}_{2,k})-\zeta_{a}(\bm{q}_{2,k}-\bm{q}_{2,k}^{\text{min}})),
	\label{eq: reflex}
\end{equation}
where $\textbf{K}^{\text{reflex}}$ is the gain of the reflex, $\bm{q}_{2,k}^{\text{max}}$ and $\bm{q}_{2,k}^{\text{min}}$ are the maximum and minimum joint angle limits, respectively, and $\zeta_{a}$ is the element-wise application of the sigmoid function with gain $a$: $\zeta_{a}(x)=1/(1+\exp(-ax))$.

Finally, the lower-layer policy is as follows
\begin{equation}
\textbf{u}^{\ddag}_{k}=\textbf{u}^{\text{DMP}}_{k}+\textbf{u}^{\text{reflex}}_{k},
	\label{eq: u low}
\end{equation}
where $\textbf{u}^{\text{DMP}}=\bm{q}_{2}^{\text{DMP}}$.
The lower-layer policy is sent to the low-level controller to compute the control input $\textbf{u}$ in (\ref{eq: u act}).

\section{Experimental settings}
\label{sec: Experimental settings}

\begin{table}[t]
  \caption{
    Movements learned in the experiment.
    This table shows the motion number, the name of the motion corresponding to the number, and the environment in which each motion is learned.
  }
  \label{tab: task}
  \begin{center}
    \begin{tabular}{|c|c|c|c|}  \hline
       Number  & Name             & Environment  \\ \hline\hline
       1       & Standing         & Flat surface \\ \hline
       2       & Swaying          & Flat surface \\ \hline
       3       & Squatting        & Flat surface \\ \hline
       4       & One-leg standing & Flat surface \\ \hline
       5       & Walking forward  & Flat surface \\ \hline
       6       & Walking backward & Flat surface \\ \hline
       7       & Sidestepping     & Flat surface \\ \hline
       8       & Jogging          & Flat surface \\ \hline
       9       & Skating          & Ramp         \\ \hline
      10       & Skating          & Jump ramp    \\ \hline
    \end{tabular}
  \end{center}
\end{table}

\subsection{Task details}

Our proposed framework was evaluated in motion learning tasks.
A total of 10 movements were learned: 8 movements on a flat surface and 2 movements on curved surfaces, as shown in Table \ref{tab: task}.
While each motion was generated $N_{\text{exp}}$ iterations, the middle and lower layer control policies were learned at each iteration, and the upper layer dynamics model was trained every 10 iterations, i.e. $N_{\text{RL}}=N_{\text{exp}}/10$.
Learning performance was evaluated using control performance and model accuracy.

The control performance at time $T$ was calculated as the sum of the cost function $\ell$:
\begin{equation}
	J_{\text{per}}(T)=\sum_{t=0}^{T}\ell(\textbf{x}_{t}, \textbf{u}_{t}).
	\label{eq: control performance}
\end{equation}
If the control performance exceeded a threshold: $J_{\text{per}}(T)>J_{\text{th}}$, motion generation was considered to be failed and terminated.
The threshold was set at $350$.
The cost function was the same as for whole-body MPC and is given by
\begin{equation}
	\ell=\|\textbf{x}_{k}-\textbf{x}_{k}^{\text{ref}}\|^{2}_{W_{\text{x}}}+\|\textbf{u}_{k}\|^{2}_{W_{\text{u}}},
	\label{eq: cost function}
\end{equation}
where $\textbf{x}^{\text{ref}}$ is the reference trajectory, and $W_{\text{x}}$ and $W_{\text{u}}$ are weights for states and input costs, respectively.
In this paper, we used the DMP trajectory $\textbf{x}^{\text{DMP}}$ as the reference.
The state weight is a diagonal matrix and consists of the weights for the generalized position and the generalized velocity.
The input weight is also a diagonal matrix and represents the weights for each joint.

The model accuracy was calculated using the $N$-step loss function in (\ref{eq: analytical model objective function}) for the observed states $(\textbf{x}_{0},\textbf{x}_{1},\textbf{x}_{2},\cdots)$ and applied inputs $(\textbf{u}_{0},\textbf{u}_{1},\textbf{u}_{2},\cdots)$ at each iteration, while the predicted state $\hat{\textbf{x}}$ in (\ref{eq: analytical model objective function}) was calculated using the augmented model: $\hat{\textbf{x}}_{k+t}=f(\hat{\textbf{x}}_{k+t-1}, \textbf{u}_{k+t-1})$ and $\hat{\textbf{x}}_{k}=\textbf{x}_{k}$.
We set the weight on the prediction error, $W_{\text{A}}$, to the same value as the state weight, $W_{\text{x}}$.

\subsection{Learning details}
\label{secsec: learning details}
The control period was different for each layer: $60$ ms for the upper layer, $30$ ms for the middle layer, and $10$ ms for the lower layer.

The analytical model is given by (\ref{eq: analytical dynamics}), and the discretization step size $\Delta t_{d}$ was set to $60$ ms.
The inertia matrix and bias forces were computed using the composite rigid body algorithm and the recursive Newton-Euler algorithm \citep{featherstone2014rigid}.
The contact forces were computed using a convex contact solver \citep{todorov2011convex}, while we approximated the friction cone as a pyramid to solve a quadratic programming problem.
To optimize the analytical model parameters $\theta_{\text{A}}$, we used a cross-entropy method \citep{kochenderfer2019algorithms}.
The initial parameters were determined from CAD data or manually.

As the residual model, we used a neural network with $2$ fully connected hidden layers of $64$ nodes with Softplus activations \citep{dugas2000incorporating} and a fully connected output layer.
The network was trained using the Adam optimizer \citep{kingma2014adam} with learning rate of $0.0001$ and batch size of $128$.
The data were pre-processed by subtracting the mean and dividing by the standard deviation.

The amplitude of the noise in the stochastic term, $\sigma$, was initially set at $0.7$ and reduced to a minimum of $0.1$ depending on the accuracy of the model.

We used iterative linear quadratic Gaussian (iLQG) \citep{todorov2005generalized} to numerically solve the optimal control problem of stochastic whole-body MPC.
The horizon was set to $N=12$ as the longest horizon that the policies could be computed within the control period of the upper layer.
Thus, MPC optimized the input sequence for $N \Delta t_{d}=0.72$  seconds.
The cost and dynamics derivatives of iLQG were computed using the finite difference method.
Their computations were parallelly performed with multiple computer threads using Open Multi-Processing (OpenMP). 
All upper layer inputs were initialized with zero and no warm start technique was used.

The middle layer inputs $\textbf{u}^{\dag}$ and gain $\textbf{K}^{\dag}$ were all initialized with zero, and state $\textbf{x}^{\dag}$ was initialized with the initial state of the robot.
To learn these values, the learning rate $\alpha=0.05$ was used, whereas $\alpha=0$ was used if a motion failed to be generated.
The forgetting rate was set to $\mu=0.01$.

The DMP trajectories for the eight movements on the flat surface were learned from motion trajectories converted from human motion data using a motion retargeting method \citep{ayusawa2017motion}.
The DMP trajectories for the skating motions were learned from the robot motion data.
The robot motion data was collected while the robot's torso was supported by a human hand.
To generate the motion data, the robot was controlled to follow a target trajectory produced from a pattern generator \citep{morimoto2008biologically}.
The reflex gain $\textbf{K}^{\text{reflex}}$ in the short latency response was a diagonal matrix with all diagonal elements equal to $0.1$.
The gain of the sigmoid function, $a$, was set to $10.0$.

\subsection{Hardware details}
All experiments were performed on a single computer with an Intel Core i9-10980XE CPU at 3.00 GHz, and we used a small-sized humanoid robot, ROBOTIS OP3.
The robot has a Dynamixel servo motor (XM-430-W350-R) at each joint.
The joint angles and angular velocities were provided by the servos.
The orientation and angular velocity of the floating base were observed using MicroStrain 3DM-GX5-AHRS IMU sensor, and the position and velocity were calculated from the sensor signals of Intel RealSense Tracking Camera T265 and the IMU sensor.
Since only the lower limbs were controlled using our proposed framework, the number of joints was $n_{j}=12$.
For proportional control of the servo motors, a gain of 70$\%$ of the default value was used.
We mounted wheels on the soles of both feet of the robot when generating the skating movements.

\begin{figure*}[t]
  \begin{minipage}[b]{0.48\linewidth}
    \centering
    \includegraphics[scale=0.64]{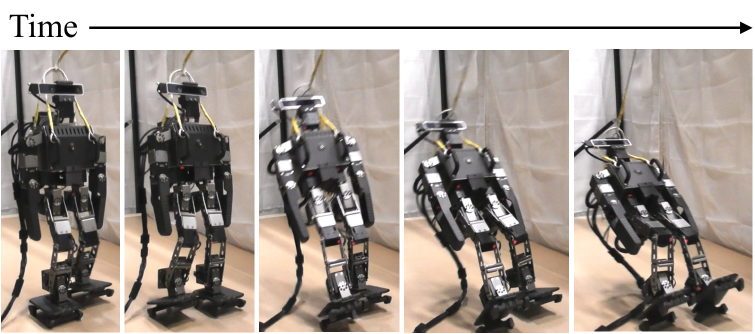}
    \subcaption{Jogging generated in the early learning phase}
    \label{fig: jog_before}
  \end{minipage}
  \begin{minipage}[b]{0.48\linewidth}
    \centering
    \includegraphics[scale=0.64]{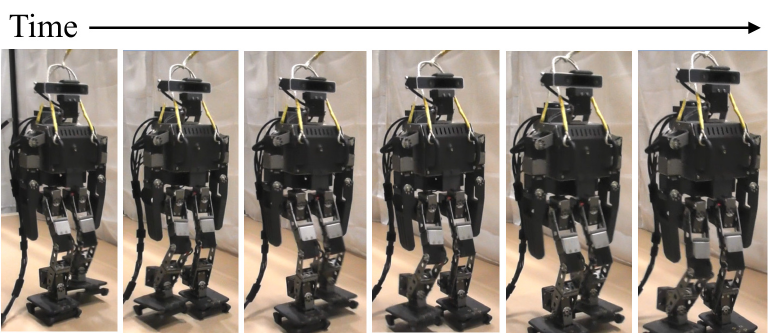}
    \subcaption{Jogging generated in the late learning phase}
    \label{fig: jog_after}
  \end{minipage}
  \caption{
    Jogging generated in (a) the early learning phase and (b) the late learning phase.
    (a) In the early learning phase, the robot fell over and could not generate a jogging, but (b) in the late learning phase, the robot successfully generated a jogging by iterative learning with the proposed framework.
  }
  \label{fig: compare learning}
\end{figure*}
\begin{figure*}[t]
  \begin{minipage}[b]{0.43\linewidth}
    \centering
    \includegraphics[scale=0.63]{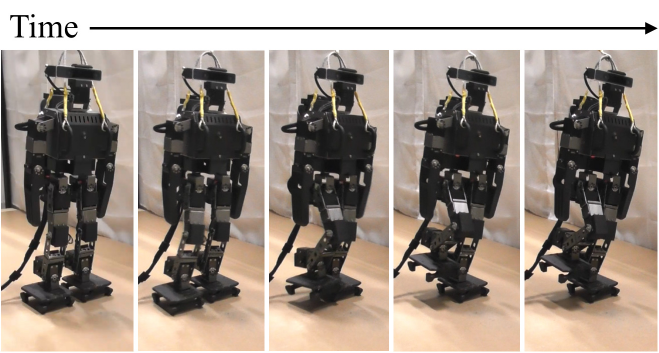}
    \subcaption{One-leg standing generated in the late learning phase}
    \label{fig: lstand_after}
  \end{minipage}
  \begin{minipage}[b]{0.53\linewidth}
    \centering
    \includegraphics[scale=0.63]{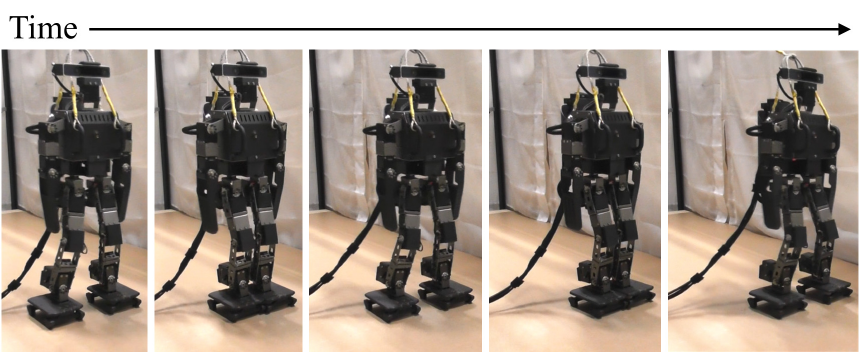}
    \subcaption{Sidestepping generated in the late learning phase}
    \label{fig: lside_after}
  \end{minipage} \\
  \begin{minipage}[b]{0.99\linewidth}
    \centering
    \includegraphics[scale=0.64]{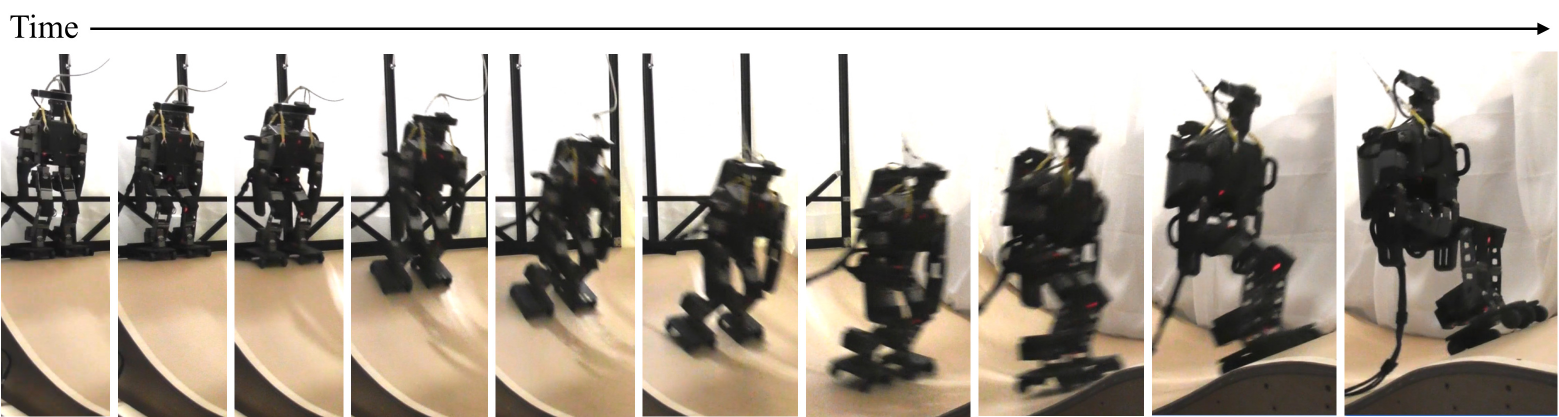}
    \subcaption{Skating on a ramp generated in the late learning phase}
    \label{fig: slideramp_after}
  \end{minipage} \\
  \begin{minipage}[b]{0.99\linewidth}
    \centering
    \includegraphics[scale=0.64]{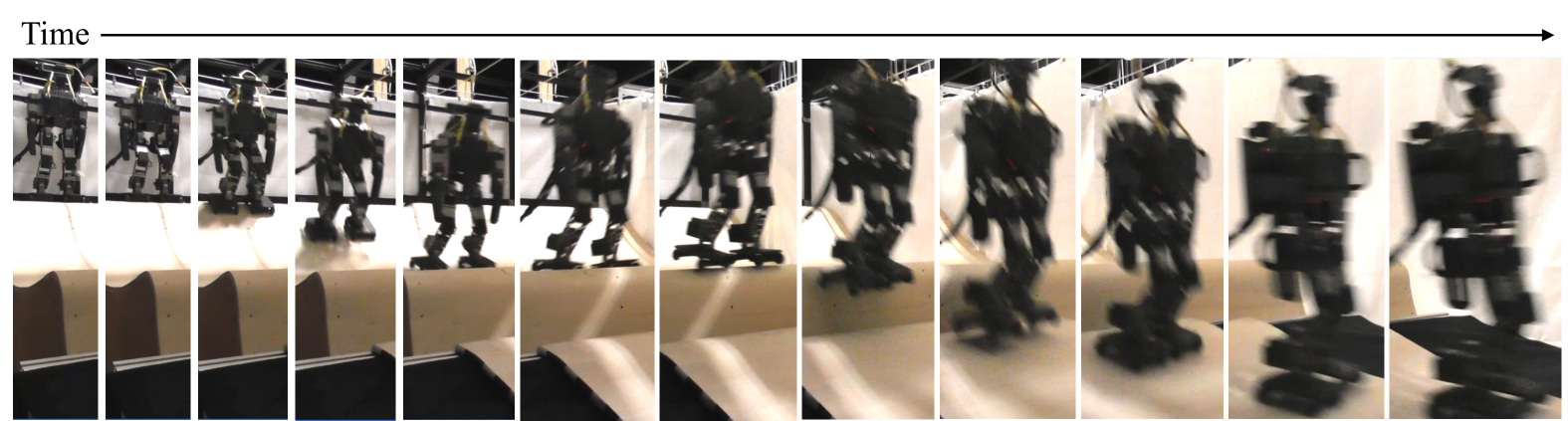}
    \subcaption{Skating on a jump ramp generated in the late learning phase}
    \label{fig: jumpramp_after}
  \end{minipage} \\
  \caption{
    Parts of other motions generated in the late learning phase: (a) one-leg standing, (b) sidestepping, (c) skating on a ramp, and (d) skating on a jump ramp.
    A variety of movements, including skating on curved surfaces, could be generated by learning with our proposed framework.  
  }
  \label{fig: versatile motion generation}
\end{figure*}

\section{Results}
\label{sec: Results}
\subsection{Learning a variety of motions}

Our proposed framework was applied to a real small humanoid robot to learn the 10 different motions listed in Table \ref{tab: task}.
By iteratively learning with our proposed framework, all motions were successfully generated in the late phases of learning.
For example, Figure \ref{fig: compare learning} shows the jogging generated in the early and late learning phases.
Figure \ref{fig: versatile motion generation} presents some of the other movements generated in the late learning phases: one-leg standing in Figure \ref{fig: lstand_after}, side stepping in Figure \ref{fig: lside_after}, and skating on a ramp and jump ramp in Figure \ref{fig: slideramp_after} and \ref{fig: jumpramp_after}, respectively.
All learned motions are shown in the accompanying video.

\begin{figure*}[t]
  \begin{minipage}[b]{0.49\linewidth}
    \centering
    \includegraphics[scale=0.175]{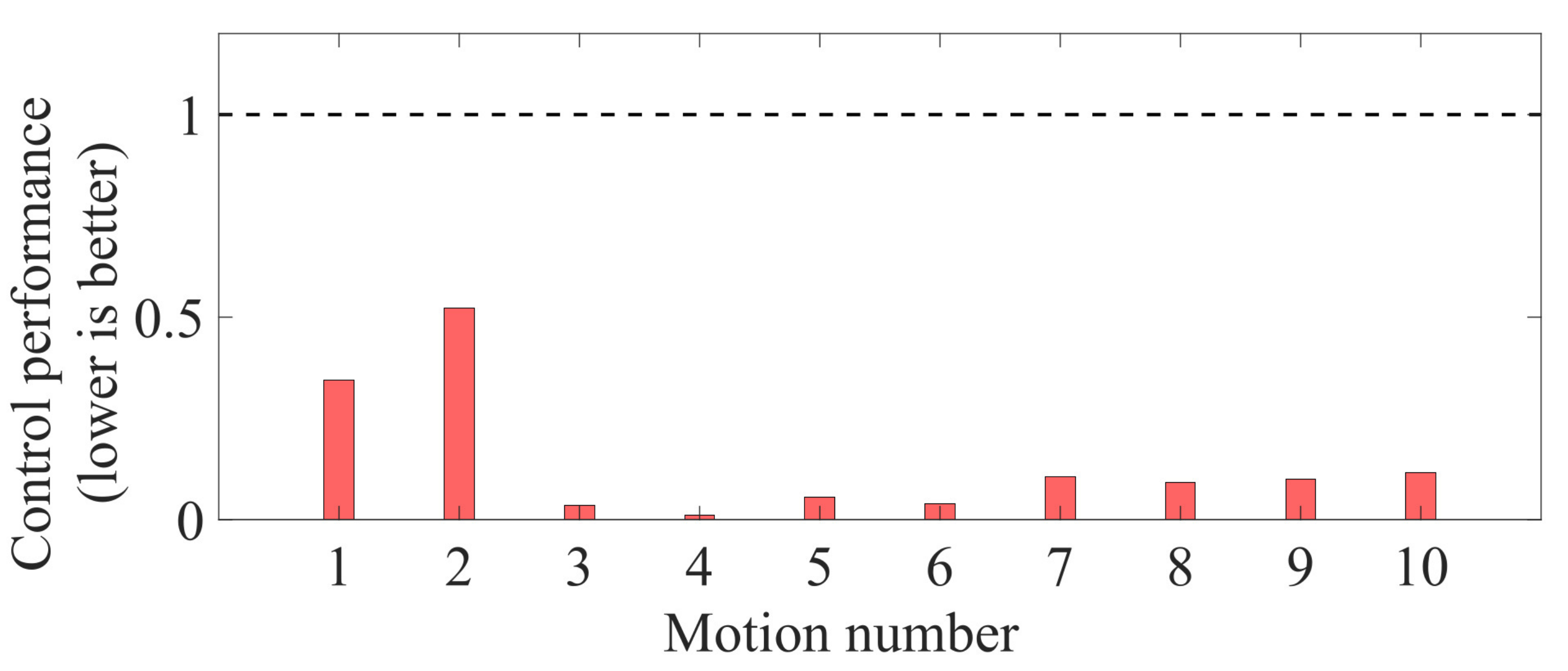}
    \subcaption{Control performance}
    \label{fig: control performance}
  \end{minipage}
  \begin{minipage}[b]{0.49\linewidth}
    \centering
    \includegraphics[scale=0.175]{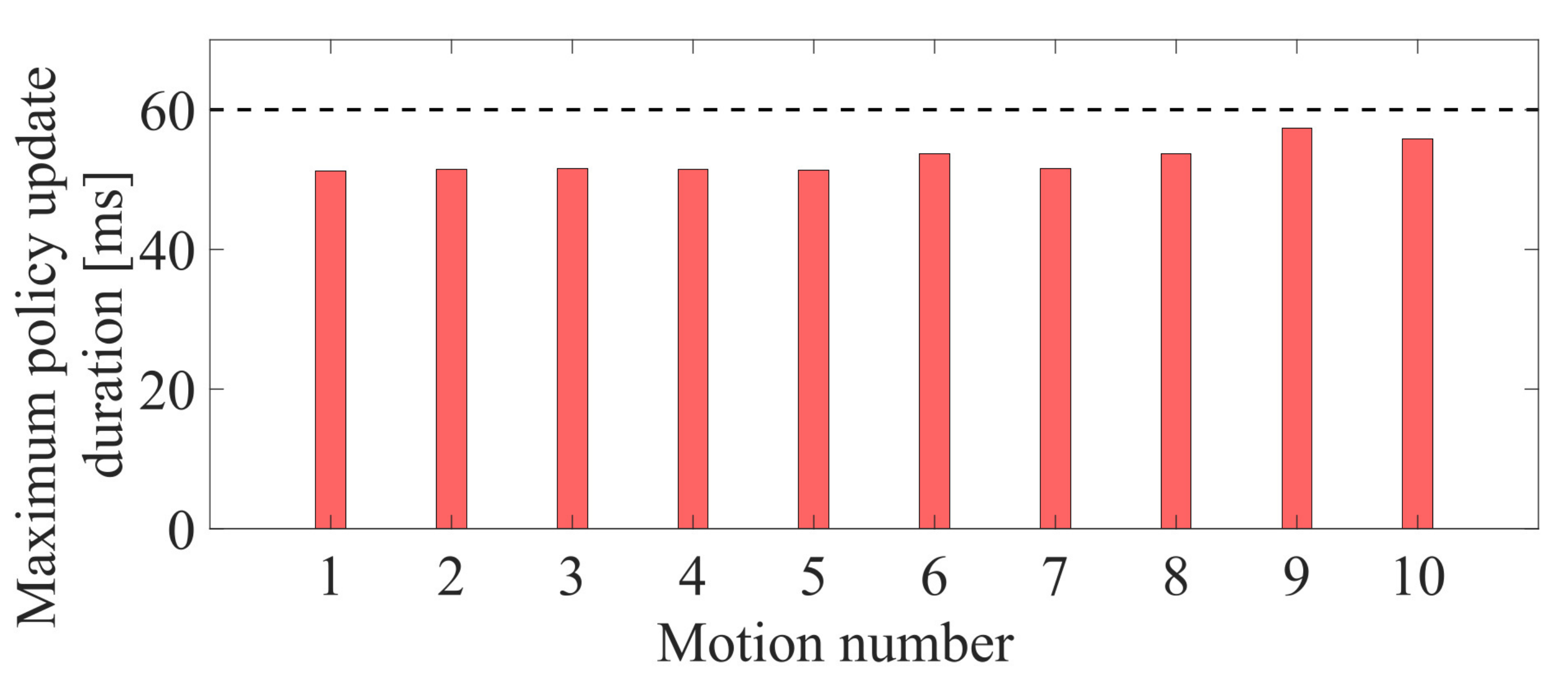}
    \subcaption{Maximum policy update duration}
    \label{fig: maximum duration}
  \end{minipage}
  \caption{
    (a) Comparison of best and initial control performance and (b) maximum update duration of the control policy for each motion.
    The numbers on the horizontal axis represent each movement, and their correspondence is shown in Table \ref{tab: task}.
    (a) The best control performances are represented by red bars and the initial control performances are represented by a black dashed line. 
    This figure represents the normalized performance, i.e., the total cost divided by the initial performance, so a smaller value indicates better performance, and the initial performance is all 1. 
    The control performances for all motions were improved by learning with our proposed framework.
    (b) The maximum update durations are indicated by red bars.
    Although updating the upper-layer policies took a long time, up to 50 ms or more, all motions were successfully generated even with such slow policy updates.
  }
  \label{fig: performance bar}
\end{figure*}

A comparison of the best and initial control performance for each learned movement is shown in Figure \ref{fig: control performance}.
The number on the horizontal axis represents each movement, e.g., 1 for standing and 10 for skating on a jump ramp (see Table \ref{tab: task} for all correspondences between numbers and movements).
The best control performances are represented by red bars, and the initial control performances are indicated by a black dashed line.
This figure shows the normalized control performance, which is the total cost of (\ref{eq: control performance}) divided by the initial control performance; smaller values indicate better performance, and the initial control performance values are all 1, as indicated by the black dashed line.
In Figure \ref{fig: control performance}, for standing (motion number: 1) and swaying (motion number: 2), the differences between the initial and best control performances are small because these motions could be generated from the initial learning phases.
The other motions were difficult to generate in the early phases of learning, and their control performances were improved by learning.
Figure \ref{fig: maximum duration} shows the maximum update duration of the control policy in the upper layer when the best control performance was obtained.
This update duration represents the maximum time between sending the observed state to the upper layer and receiving the control input from the upper layer, which includes the computation time of whole-body MPC.
Figure \ref{fig: maximum duration} shows that the upper layer takes a maximum update duration of about $50$ ms, with the longest update duration being $57.35$ ms for skating on the ramp (motion number: 9).
Even with such long update durations, our proposed framework allowed to generate robot motions using whole-body MPC.
The control period for the upper layer is $60$ ms, which is plotted as a dashed black line in Figure \ref{fig: maximum duration}.
All maximum update times were shorter than the control period, thus whole-body MPC was able to generate all movements in real time.

\subsection{Learning curves for control performance and model accuracy}

\begin{figure*}[t]
  \begin{minipage}[b]{0.32\linewidth}
      \centering
      \includegraphics[scale=0.175]{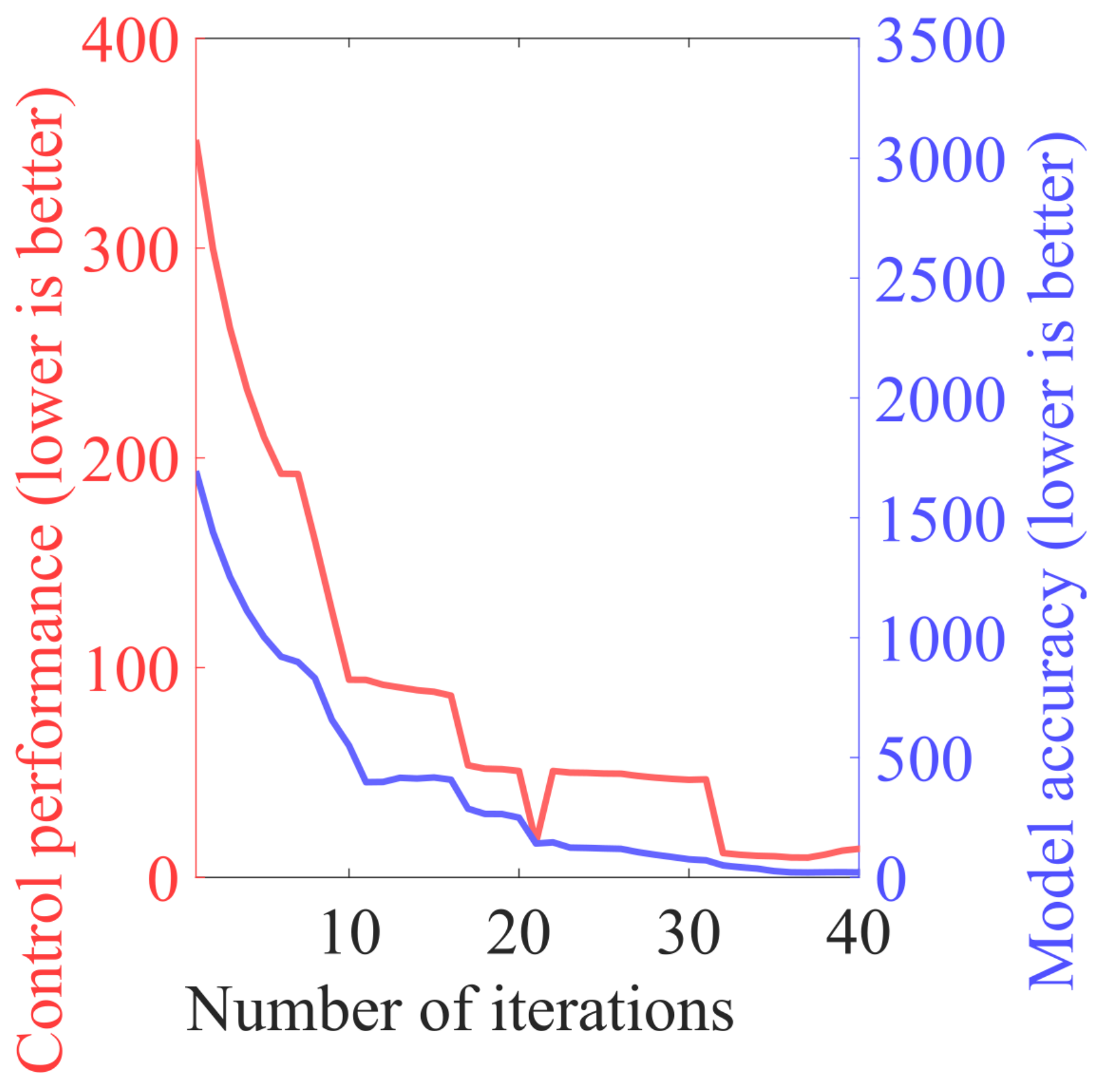}
      \subcaption{One-leg standing}
      \label{lcurve_lstand}    
  \end{minipage} 
  \begin{minipage}[b]{0.32\linewidth}
    \centering
    \includegraphics[scale=0.175]{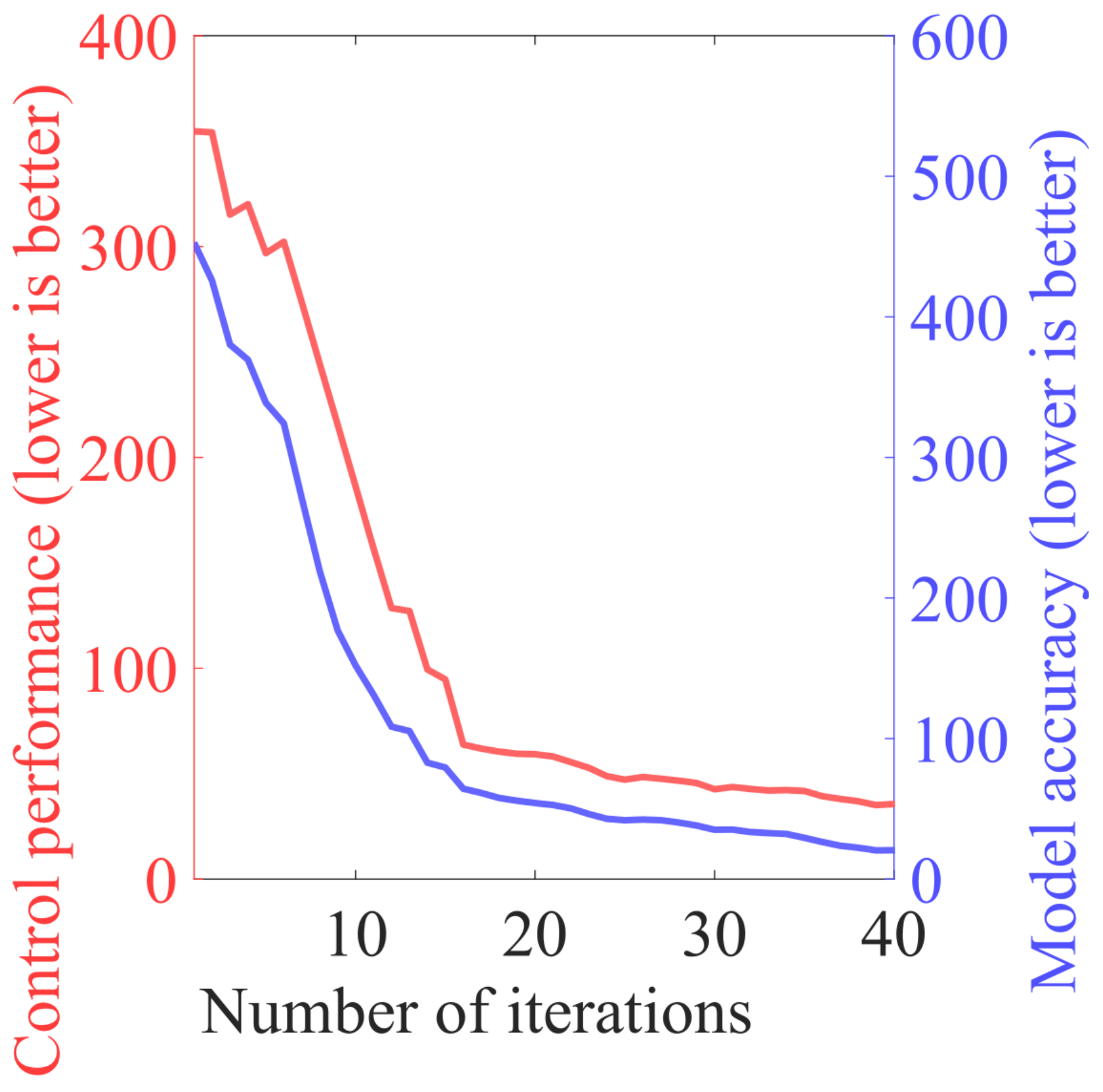}
    \subcaption{Walking forward}
    \label{lcurve_walk}    
  \end{minipage} 
  \begin{minipage}[b]{0.32\linewidth}
    \centering
    \includegraphics[scale=0.175]{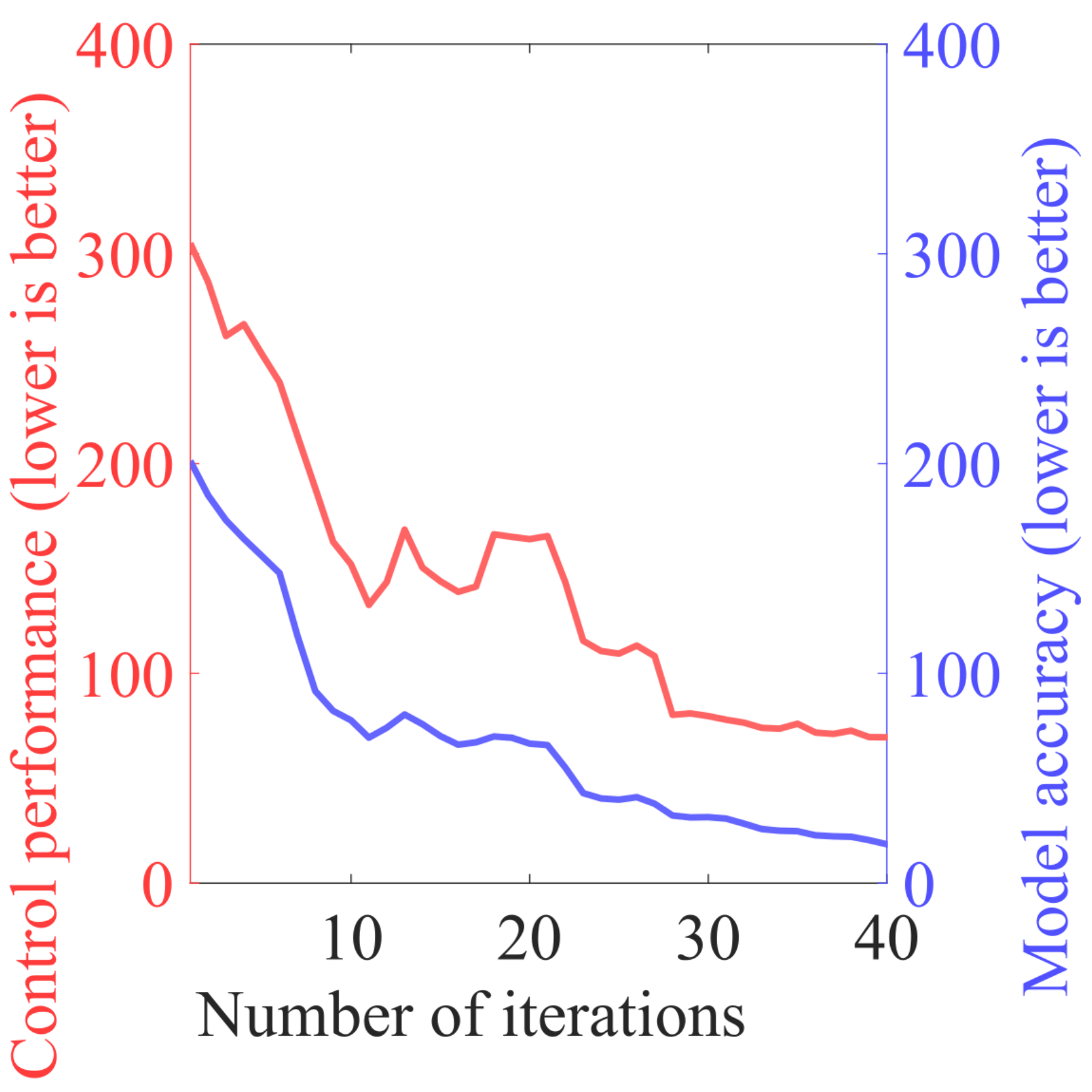}
    \subcaption{Sidestepping}
    \label{lcurve_lside}    
  \end{minipage} \\  
  \begin{minipage}[b]{0.32\linewidth}
    \centering
    \includegraphics[scale=0.175]{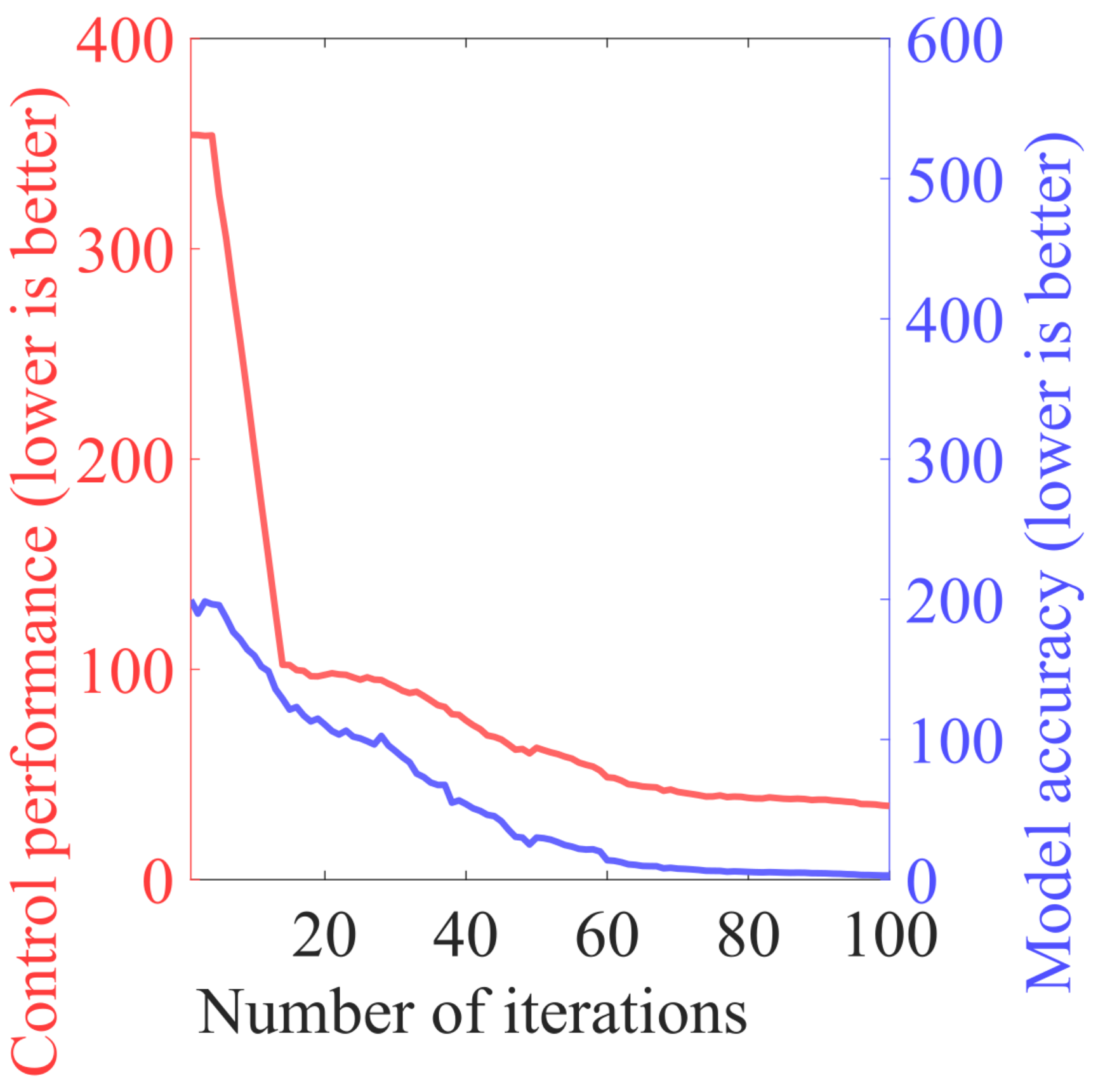}
    \subcaption{Jogging}
    \label{lcurve_jog}
  \end{minipage} 
  \begin{minipage}[b]{0.32\linewidth}
    \centering
    \includegraphics[scale=0.175]{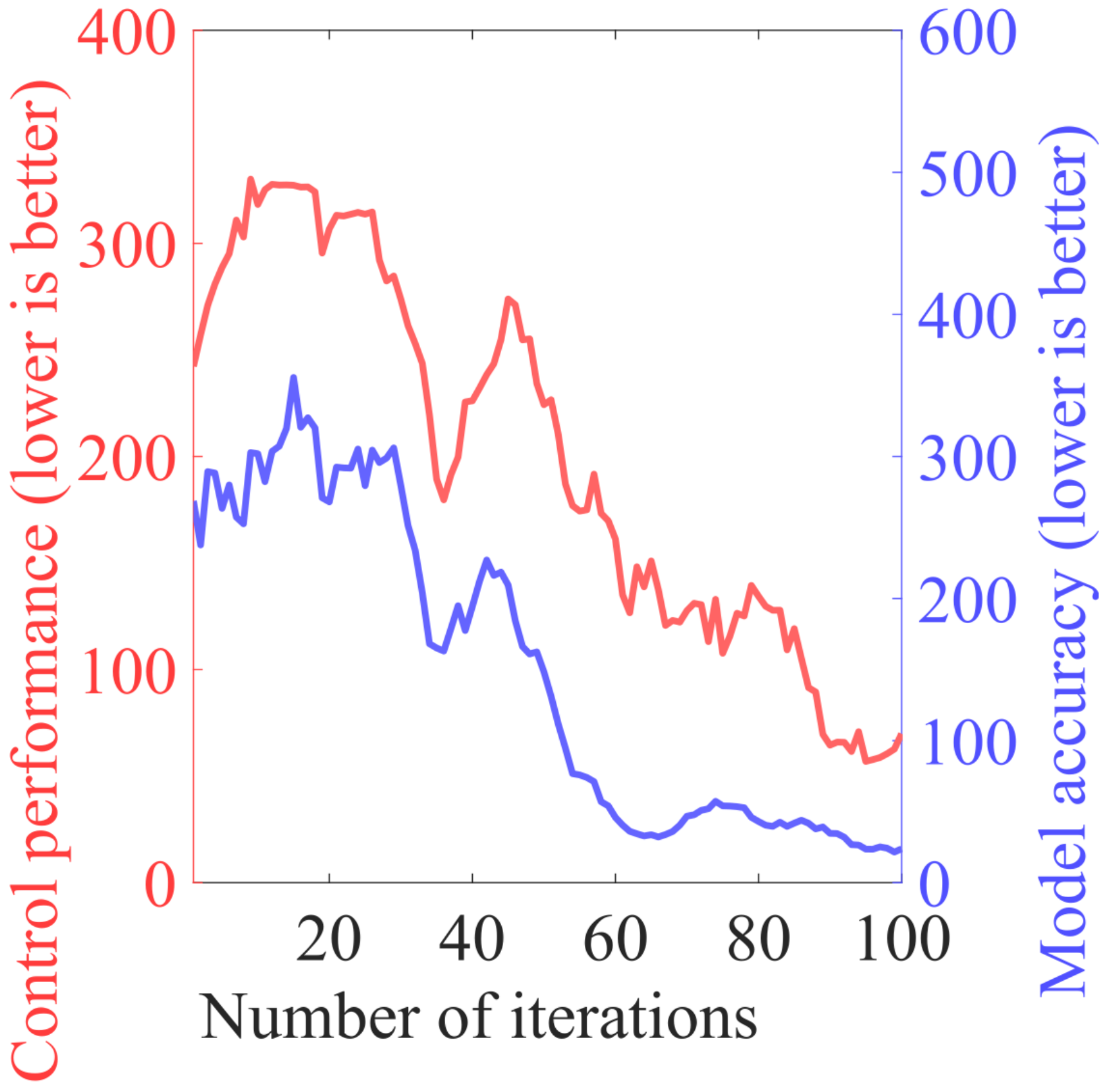}
    \subcaption{Skating on a ramp }
    \label{lcurve_slideramp}
  \end{minipage} 
  \begin{minipage}[b]{0.32\linewidth}
    \centering
    \includegraphics[scale=0.175]{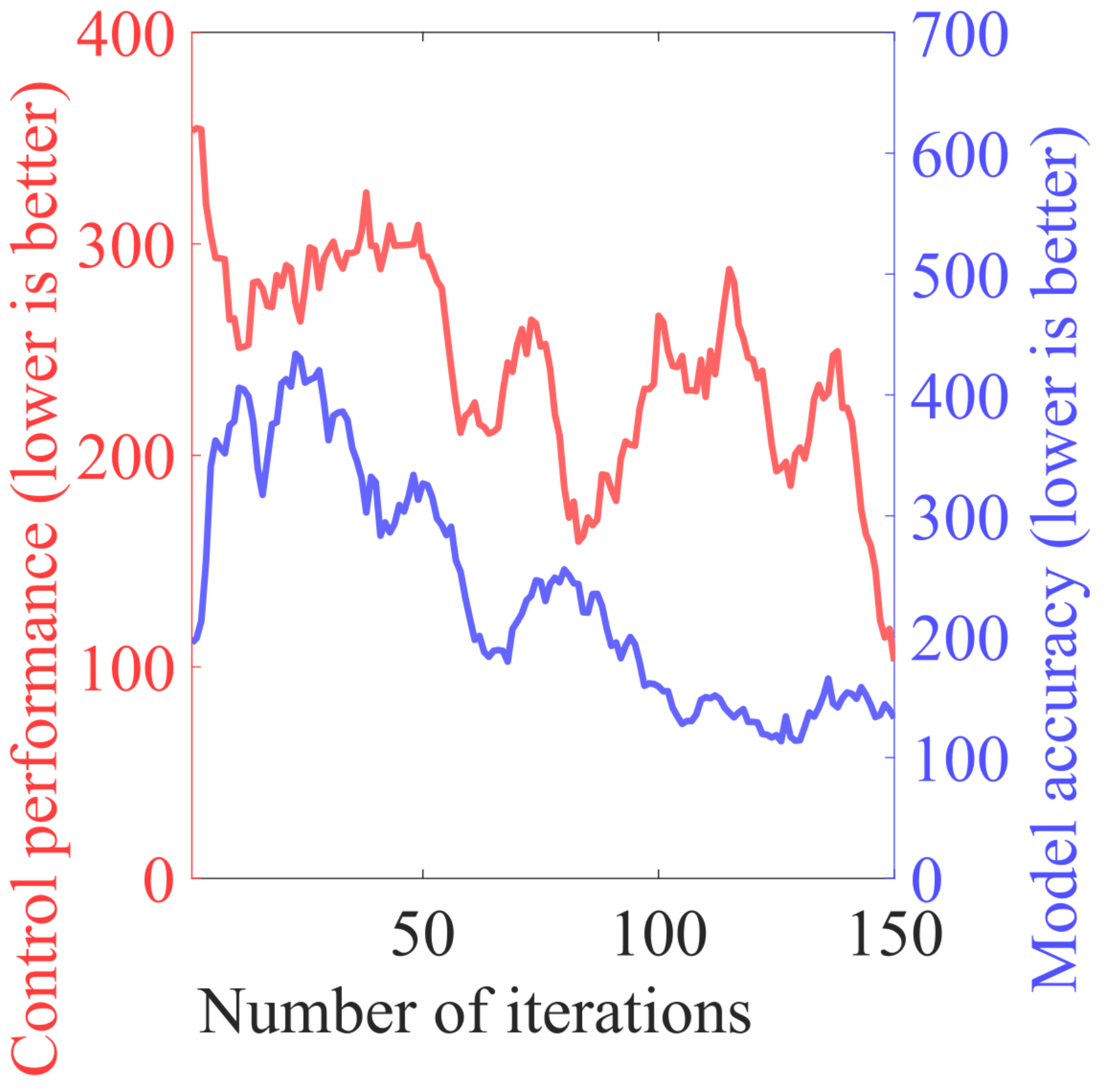}
    \subcaption{Skating on a jump ramp}
    \label{lcurve_jumpramp}
  \end{minipage} \\
  \caption{
    Learning curves plotting control performance and model accuracy for (a) one-leg standing, (b) walking forward, (c) sidestepping, (d) jogging, (e) skating on a ramp, and (f) skating on a jump ramp. 
    The red and blue lines are 10-point moving averages of control performance and model accuracy, respectively.
    Control performance was calculated as the sum of costs, and model accuracy was evaluated using an N-step loss function, with smaller values indicating better performance and accuracy. 
    The control performance was low in the early learning phase, but as the iterations progressed, the control performance improved, and in the late learning phase, all motions could be generated successfully.
    In addition, the accuracy of the model tended to improve as the learning progressed, and in the later phases of learning, accurate models were acquired that allowed whole-body MPC to generate a wide variety of robot motions.    
  }
  \label{fig: learning curve}
\end{figure*}

Figures \ref{lcurve_lstand} - \ref{lcurve_jumpramp} respectively represent the learning curves for one-leg standing, forward walking, sidestepping, jogging, skating on a ramp, and skating on a jump ramp.
In these learning curves, the left vertical axis represents control performance and the horizontal axis represents the number of learning iterations.
The control performances are not normalized in this figure.
Each movement was learned only once because the robot was prone to breakage, especially in agile movements such as skating on a jump ramp.
The red line represents a moving average of the 10 data points of control performance to show trends.
As can be seen from Figure \ref{fig: learning curve}, in the early learning phase, the control performance often exceeded the threshold $J_{\text{th}}=350$, i.e., the robot failed to generate those motions, but the iterative learning with our proposed framework tended to gradually improve the control performance as the learning progressed.
In the later phases of learning, the robot was able to generate all the motions using our proposed framework.
Note that the learning of skating on the jump ramp did not converge even in the later phases, and more iterations may be needed.

In our proposed framework, we used an uncertainty-aware model augmented with a residual network to stably learn an accurate dynamics model via model-based RL.
Figure \ref{fig: learning curve} also shows the learning curve of the dynamics model, where the right vertical axis shows the accuracy of the model.
The blue line represents a moving average of 10 data points of model accuracy to show trends.
The accuracy of the model is evaluated as an $N$-step loss function of (\ref{eq: analytical model objective function}), with lower values representing better accuracy.
As the learning iterations progress, the $N$-step loss function tended to decrease gradually and the accuracy of the model was improved.
For example, for the jogging in Figure \ref{lcurve_jog}, the initial $N$-step loss was $249.81$, which decreased to $2.43$ at the 100th iteration.
Using the learned dynamics models, a variety of real humanoid robot motions were successfully generated by whole-body MPC, thus it appears that the dynamics of the humanoid robot could be accurately learned using our augmented model.

\subsection{Ablation studies}

We conducted two ablation studies in which we removed the main components of our framework and observed how they affected control performance and model accuracy.

\begin{figure*}[t]
  \begin{minipage}[b]{0.49\linewidth}
    \centering
    \includegraphics[scale=0.17]{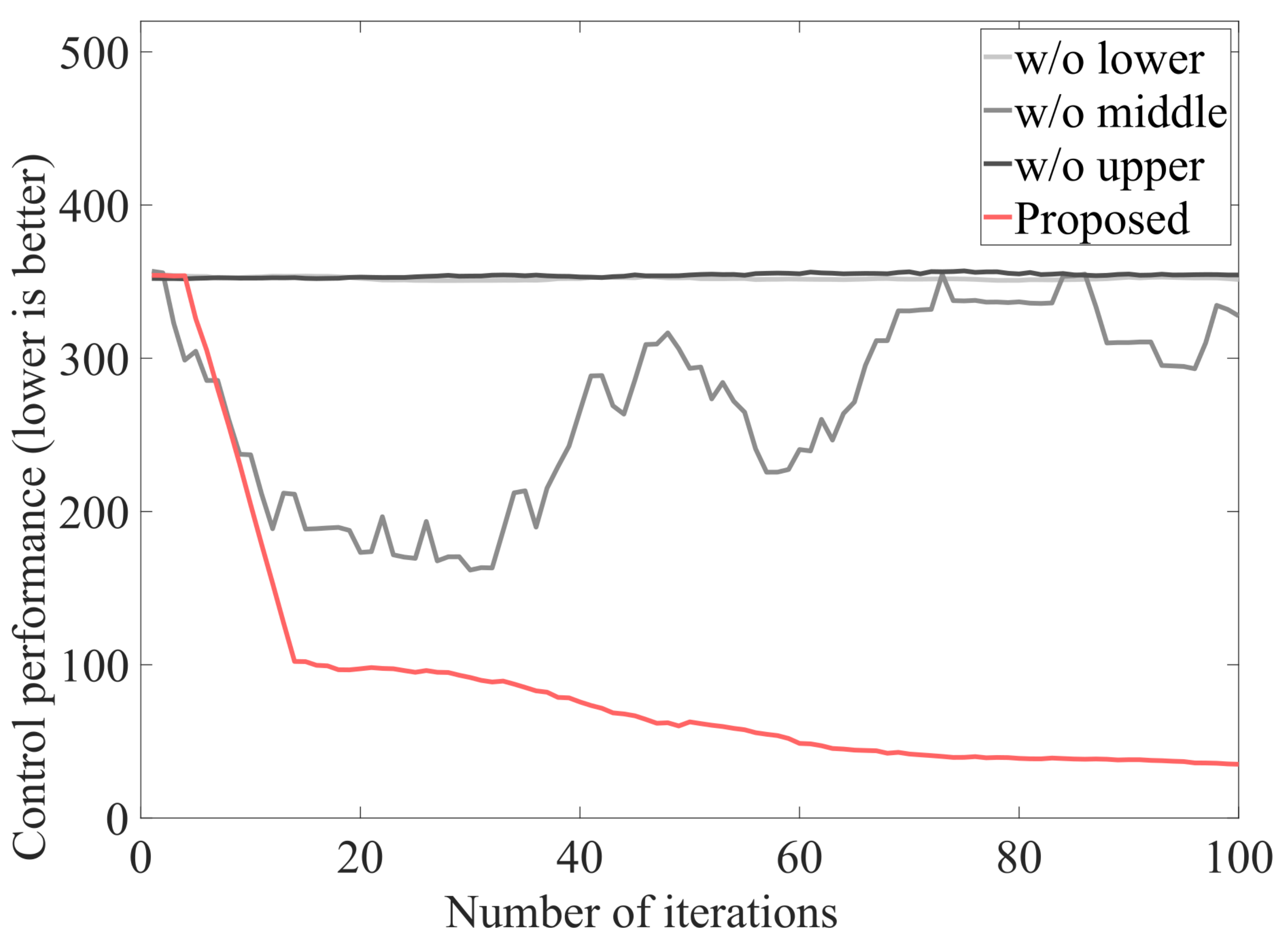}
    \subcaption{Performance in the first ablation study}
    \label{fig: ablation hierarchical}
  \end{minipage}
  \begin{minipage}[b]{0.49\linewidth}
    \centering
    \includegraphics[scale=0.17]{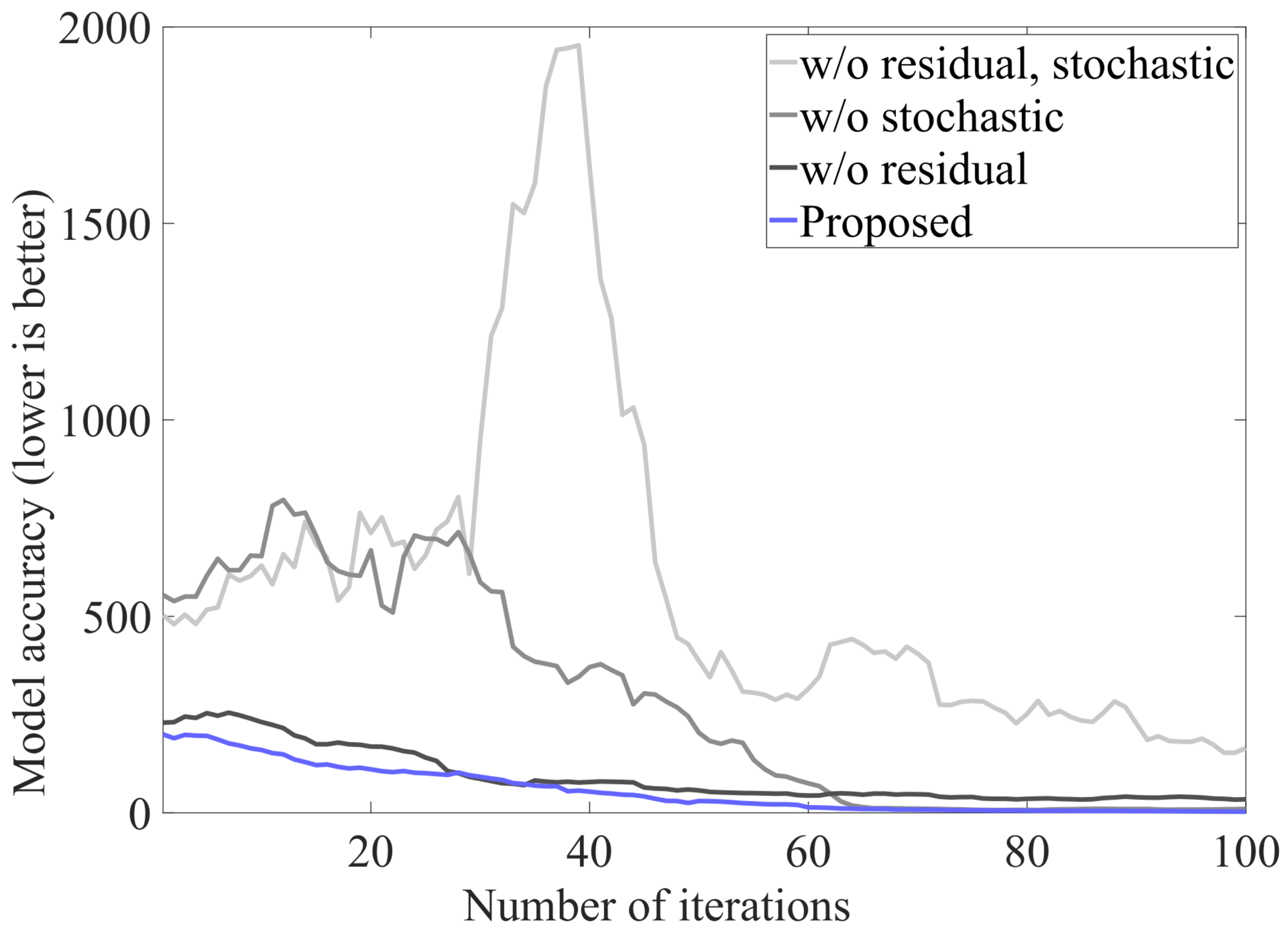}
    \subcaption{Accuracy in the second ablation study}
    \label{fig: accuracy ablation model}
  \end{minipage}  
  \caption{
    (a) Control performance in the ablation study for the hierarchy, and (b) model accuracy in the ablation study for the dynamics model.
    In these figures, moving averages are plotted, with smaller values representing better performance and accuracy.
    (a) Learning performance deteriorates when each layer was removed; therefore, all layers are considered necessary for learning.
    (b) When the model has both a residual network and a stochastic term, an accurate model was stably learned.
  }
  \label{fig: ablation study}
\end{figure*}

The first ablation study evaluated the control performance when each layer was removed from the proposed framework.
Figure \ref{fig: ablation hierarchical} shows their learning curves in learning jogging.
The robot was unable to generate jogging when the lower layer was removed (w/o lower in Figure \ref{fig: ablation hierarchical}) and when the upper layer was removed (w/o upper in Figure \ref{fig: ablation hierarchical}), even as the learning iterations progressed.
When the middle layer was removed (w/o middle in Figure \ref{fig: ablation hierarchical}), the jogging could not be learned stably and the control performance was worse than in our proposed framework (Proposed in Figure \ref{fig: ablation hierarchical}).
Therefore, all layers of our proposed framework are considered essential for learning.

\begin{figure*}[t]
  \begin{center}
    \includegraphics[width=0.99\linewidth]{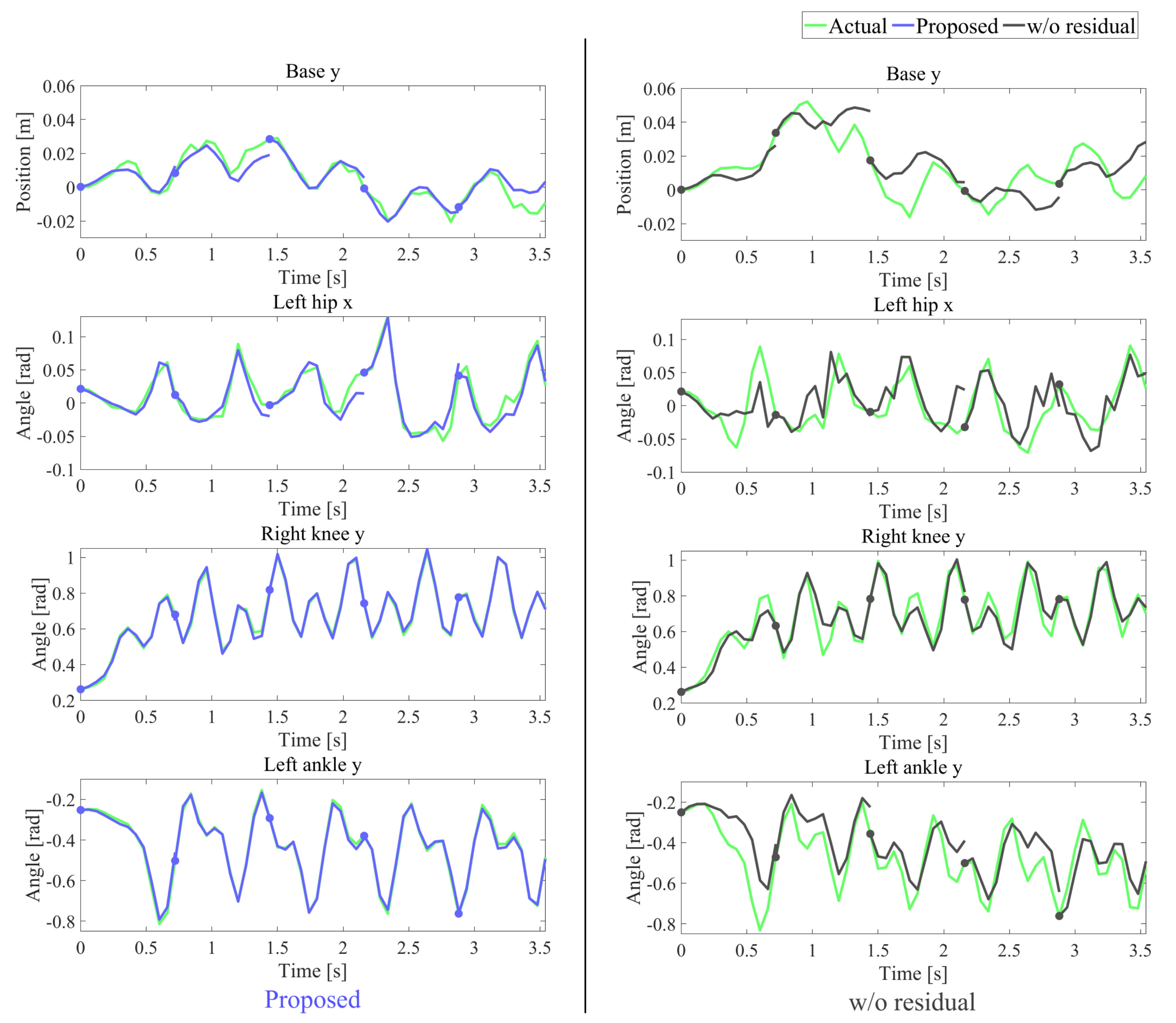}
    \caption{
      Predicted trajectories of base link position and joint angles using our augmented model and the dynamics model without the residual network for the jogging trajectory.
      Both dynamics models were trained with 100 learning iterations.
      The green lines represent the observed actual motion data, and the blue and black lines represent the trajectories predicted at intervals of $0.72$ s from $0$ s using our augmented model and the dynamics model without the residual network, respectively.
      Our augmented model was able to predict trajectories closer to the actual motion data than the dynamics model without the residual network, and thus a more accurate dynamics model was obtained by learning.
    }
    \label{fig: predicted trajectories}
  \end{center}
\end{figure*}

The second ablation study evaluated the accuracy of the model when residual network in (\ref{eq: residual model}) and the stochastic term in (\ref{eq: stochastic whole-body dynamics}) were removed from the dynamics model.
Figure \ref{fig: accuracy ablation model} shows their learning curves for learning jogging.
When the stochastic term was removed, the learning tends to be less stable, and the accuracy of the model tends to be worse, especially in the early phases of learning.
In particular, when both the residual network and the stochastic term were removed (w/o residual, stochastic in Figure \ref{fig: accuracy ablation model}), the learning is unstable, with sudden decreases and increases in model accuracy.
Therefore, the stochastic term is considered to contribute to the stability of the learning.
When the residual network was removed from the dynamics model (w/o residual in Figure \ref{fig: accuracy ablation model}), the model was less accurate than our augmented model (Proposed in Figure \ref{fig: accuracy ablation model}).
To show that the improvement in accuracy due to our augmented model is not trivial, Figure \ref{fig: predicted trajectories} shows the motion trajectories predicted by our augmented model (Proposed in Figure \ref{fig: accuracy ablation model}) and by the dynamics model without the residual network (w/o residual in Figure \ref{fig: accuracy ablation model}) after 100 learning iterations for the jogging.
In the left-hand side of Figure \ref{fig: predicted trajectories}, the actual motion data (green lines) are compared with the predicted trajectories of our augmented model (blue lines), and in the right-hand side, the actual data (green lines) are compared with the predicted trajectories of the dynamics model without the residual network (black lines).
In this figure, the trajectories of the lateral position of the base link (Base y) and the joint angles of left hip adduction/abduction (Left hip x), right knee flexion/extension (Right knee y)), and left ankle dorsiflexion/plantarflexion (Left Ankle y) were predicted for intervals of $N\Delta t_{d} = 0.72$ s, from $0$ s.
Each dot on the predicted trajectory represents the starting position or angles of the prediction.
Compared to the model without the residual network, our augmented model predicted motion trajectories closer to the actual data, indicating that more accurate dynamics could be learned using our augmented model.
Therefore, both the residual network and the stochastic term are considered necessary for accurately and stably learning the dynamics model.

\section{Conclusions}
\label{sec: Conclusions}
In this paper, we proposed a hierarchical learning framework for whole-body MPC inspired by human motor control.
In the upper layer, accurate dynamics models were learned through model-based RL with stochastic whole-body MPC to address the simulation-to-real gap problem.
The middle and lower layers additionally learned control policies that could respond faster than the upper layer, such as fast modulation of reflexes \citep{capaday1986amplitude}, so that the robot's motion could be generated even if the policy updates of the upper layer whole-body MPC were slow.

Through iterative learning with our proposed framework, the real humanoid robot successfully generated 10 different motions using whole-body MPC.
Furthermore, our proposed framework was thoroughly investigated through the ablation studies.

One of our future works is to add model-free RL to our hierarchical framework to learn control policies that achieve better control performance, especially for agile motions such as skating on a jump ramp.
For example, \cite{nagabandi2018neural} showed that control performance can be improved by initializing control policies with model-based policies and further learning with model-free RL.
Another of our future works is to develop a humanoid robot with viscoelastic properties at each joint and apply our hierarchical learning framework.
A popular hypothesis in human motor control suggests that the intrinsic viscoelastic properties of muscle mitigate the effects of slow policy updates of the central nervous system \citep{hogan1984adaptive,loeb1999hierarchical,burdet2001central}.
A humanoid robot with such viscoelastic properties would be able to spend more computation time on whole-body MPC, allowing trajectory optimization over longer horizons for better control performance.

\begin{dci}
The author(s) declared no potential conflicts of interest with respect to the research, authorship, and/or publication of this article.
\end{dci}
\begin{funding}
  The author(s) disclosed receipt of the following financial support for the research, authorship, and/or publication of this article: 
  This article is based on results obtained from a project, JPNP20006, commissioned by the New Energy and Industrial Technology Development Organization (NEDO).
  Part of this work was supported by JSPS KAKENHI Grant Number JP16H06565 and MIC-SCOPE (182107105).
\end{funding}

\bibliographystyle{SageH}
\bibliography{references.bib}

\end{document}